%% file: main.tex
\documentclass[journal, twoside]{./style/IEEEtran} 

\usepackage{subcaption}
\usepackage[numbers]{natbib}
\usepackage{multicol}
\usepackage[bookmarks=true]{hyperref}

\usepackage{relsize}
\selectfont

\newenvironment{nodiff}{}{}

\input{preamble}
\input{style/dlr_colors}

\usepackage{adjustbox}
\newlength{\bioimgw}
\newlength{\bioimgh}

\usepackage[separate-uncertainty]{siunitx}
\usepackage{bm}
\usepackage{xfrac}
\usepackage{mathtools}
\DeclarePairedDelimiterX{\norm}[1]{\lVert}{\rVert}{#1}
\newcommand{\Exp}{\mathrm{Exp}}
\newcommand{\Log}{\mathrm{Log}}

\usepackage{algorithm}
\usepackage{algpseudocode}

\usepackage[acronym]{glossaries}
\glsdisablehyper
\newacronym{sts}{STS}{single tangent space}
\newacronym{gmm}{GMM}{Gaussian Mixture Model}
\newacronym{gmr}{GMR}{Gaussian Mixture Regression}
\newacronym[longplural=Gaussian Processes]{gp}{GP}{Gaussian Process}
\newacronym{kmp}{KMP}{Kernelized Movement Primitive}
\newacronym{rbf}{RBF}{Radial Basis Function}
\newacronym[longplural=degrees of freedom]{dof}{DoF}{degree of freedom}
\newacronym{ml}{ML}{Machine Learning}
\newacronym{lfd}{LfD}{learning from demonstration}
\newacronym{ds}{DS}{Dynamical System}
\newacronym{poe}{PoE}{product of experts}
\newacronym{kl}{KL}{Kullback-Leibler}

\newcommand{\rmcaffiliation}{German Aerospace Center (DLR), Robotics and Mechatronics Center (RMC), M\"unchener Str. 20, 82234 We\ss ling, Germany.}
\newcommand{\tumaffiliation}{Department of Computer Engineering, Technical University of Munich, Friedrich-Ludwig-Bauer-Str. 3, Garching, Germany.}
\newcommand{\leedsaffiliation}{School of Computer Science, University of Leeds, Leeds, UK.}

\IEEEoverridecommandlockouts                             

\title{Fast Non-Parametric Heteroscedastic Imitation Learning With Geometric Priors}

\author{Maximilian Mühlbauer$^{1,2}$, Arne Sachtler$^{1,2}$, Markus Knauer$^{2,1}$, Cem Küçükgenç$^{1,2}$, Yanlong Huang$^3$,\\ Alin Albu-Schäffer$^{2, 1}$, João Silvério$^2$
\thanks{$^1$ \tumaffiliation}  
\thanks{$^2$ \rmcaffiliation}  
\thanks{$^3$ \leedsaffiliation}  
}

\makeatletter
\AddToHook{cmd/appendices/before}{\def\cref@section@alias{appendix}\def\cref@subsection@alias{appendix}\def\cref@subsubsection@alias{appendix}}
\makeatother

\begin{document}

\maketitle


\begin{abstract}
When learning probabilistic policies from human demonstrations, data-efficient learning and fast adaptations to new scenarios are key requirements.
One popular way to achieve intuitive and reliable adaptations is through non-parametric, typically kernel-based, methods.
However, existing solutions either fail to account for the geometry of manifolds common in robotics, limiting data efficiency, or, when geometry-aware, provide unreliable uncertainty estimates or require retraining to adapt.
We propose a non-parametric approach leveraging geometric priors in scenarios of data scarcity and heteroscedastic uncertainties for probabilistic modeling.
We utilize the method to formulate policies based on time or robot state, where non-separable diagonal kernels allow capturing uncertainty relations between degrees of freedom for same-sized in- and outputs.
Fast updates, requiring less than \SI{3}{\milli\second} for a trajectory involving both position and orientation are possible through an optimized formulation.
Our approach supports both manifold-valued input and manifold-valued output with large orientation changes.
Using task parameterization, adaptation to different object poses is easily possible.
We evaluate the approach on a set of toy examples and on real robot manipulation tasks both in autonomous execution and in shared control scenarios.
\end{abstract}


\section{Introduction}
Probabilistic representations of movement primitives are key in applications with requirements in robustness and safety guarantees, as for example in space robotics~\cite{muehlbauer2025control}, underwater robotics~\cite{schmidt2025conceptual} or healthcare assistance~\cite{quere2024probabilistic}.
\textit{Parametric} approaches such as \glspl{gmm}~\cite{zeestraten2017manifold,calinon2020gaussians} provide an \textit{aleatoric} (inherent to the data) uncertainty estimate while \textit{non-parametric} approaches like \glspl{gp}~\cite{williams2006gaussian} provide \textit{epistemic} (lack of data) uncertainty \cite{huellermeier2021aleatoric}.
\glspl{kmp}~\cite{huang2019kernelized} and their extensions~\cite{huang2021toward,silverio2023nonparametric} have shown remarkable qualities regarding modeling capability and adaptability, while in parallel providing \textit{both} uncertainty estimates.
Both uncertainty types are required in applications  -- the epistemic uncertainty estimate can, e.g., be used to render the robot compliant when unsure about its actions \cite{knauer2025interactive} while the aleatoric uncertainty helps to compute control gains encoding task variability \cite{muehlbauer2025unified}.
Heteroscedasticity allows our method to model different uncertainties for different phases of a task, which is a key requirement in policy arbitration \cite{muehlbauer2024probabilistic,muehlbauer2025unified}.
\glspl{kmp} have already been applied to a range of learning problems, such as manufacturing tasks~\cite{knauer2025interactive}, in space applications~\cite{muehlbauer2025control} or to guide reinforcement learning~\cite{padalkar2025towards}.

\glsreset{sts}
Currently, these approaches are however only designed for Euclidean~\cite{huang2019kernelized,silverio2023nonparametric} or \gls{sts}~\cite{huang2021toward}, linearizing curved manifolds at one point, thus inducing distortions~\cite{jaquier2024unraveling} and not allowing to incorporate prior geometric knowledge.
In contrast, approaches for multiple tangent spaces as already developed for \glspl{gmm}~\cite{zeestraten2017manifold} induce much less distortion through the utilization of multiple linearization points.
A notable exception for \glspl{kmp} is developed in~\cite{li2025orientation} fusing multiple \gls{sts} \glspl{kmp}~\cite{huang2021toward} on multiple tangent spaces on $\mathrm{SO}(3)$, at the cost, however, of an intricate formulation and a lack of generalizability to other geometries.
Additionally, each of the \gls{sts} \glspl{kmp} still suffers from issues resulting from using a single linearization point.
Conversely, \citet{lopez2023non} developed a kernel-based, on-manifold approach which, however, uses different distributions per manifold which are hard to join in product manifolds and computationally slow.
This approach also only outputs aleatoric uncertainty, neglecting epistemic uncertainty which is fundamental in many imitation learning problems.
Our solution (\Cref{fig:intro:overview}) combines the advantages of the aforementioned methods, namely fast predictions with aleatoric and epistemic uncertainty.
On selected manifolds such as $\mathrm{SO}(3)$, we achieve global validity through the double-coverage by $\mathcal{S}^3$.
Using the Riemannian Gaussian distribution~\cite{zeestraten2017manifold,calinon2020gaussians}, no discrete linearization points are required, removing the covariance distortions caused by \gls{sts} methods (\Cref{sec:kmp}).

\ifdefined\rkmpu\else\newlength{\rkmpu}\fi

\begin{figure}
  \centering
  \resizebox{\columnwidth}{!}{
  \setlength{\rkmpu}{0.0768\textwidth}
  \begin{tikzpicture}[
      x=\rkmpu, y=\rkmpu,
      head/.style     = {font=\sffamily\large\bfseries},
      section/.style  = {font=\sffamily\small\bfseries, anchor=west},
      boxtitle/.style = {font=\sffamily\small},
      secref/.style   = {font=\sffamily\small},
      img/.style      = {inner sep=0pt},
      outerbox/.style = {draw=dlrgreen1,  line width=1.5pt},
      orangebox/.style= {draw=dlrblue1, line width=1.2pt},
    ]

    \draw[outerbox] (0.18,-0.33) rectangle (13.08,-6.52);
    \fill[dlryellow3] (7.67,-1.13) rectangle (12.85,-6.32);   
    \fill[dlrgreen5]  (3.97,-1.23) rectangle (7.45,-3.45);    
    \fill[pastelpurple]  (3.97,-3.80) rectangle (7.45,-6.12);    
    \fill[dlrblue5]  (7.85,-3.80) rectangle (12.75,-6.12);   

    \node[head] at (6.60,-0.65) {Riemannian Kernelized Movement Primitives};
    \node[secref, anchor=east] at (12.95,-0.65) {\ref{sec:kmp:prediction}};

    \node[section] at (0.50,-1.42) {State-based Policies};
    \node[img] at (2.10,-2.60)
      {\includegraphics[width=3.10\rkmpu,height=1.70\rkmpu,keepaspectratio,page=9, trim={20 0 20 50},clip]{bilder}};

    \node[boxtitle] at (5.70,-1.47) {Matrix-valued Kernels};
    \node[secref, anchor=north east] at (7.43,-1.26) {\ref{sec:matrix_kernels}};
    \node[img] at (5.70,-2.53)
      {\includegraphics[width=3\rkmpu,height=1.7\rkmpu,keepaspectratio, trim={0 70 0 70},clip]{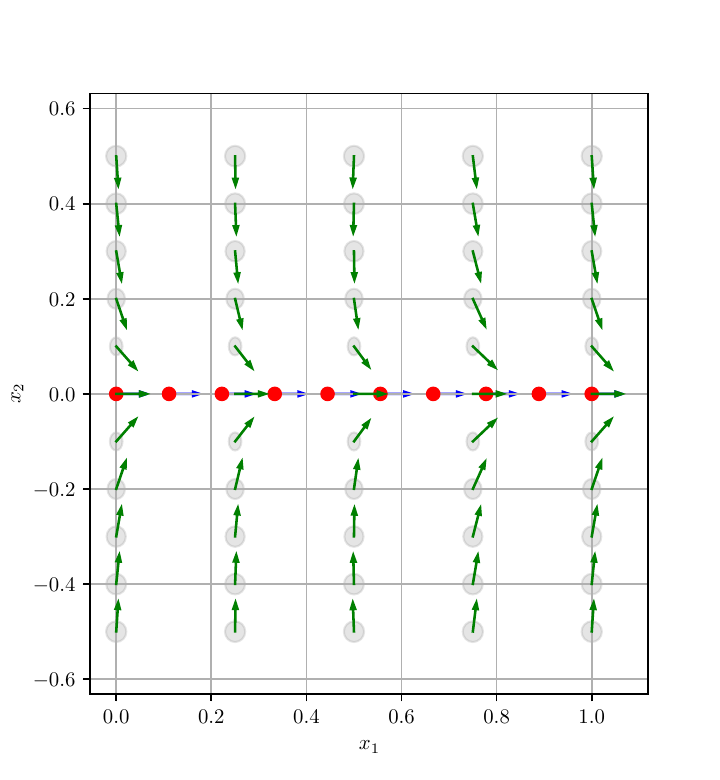}};

    \node[fill=white, rotate=10, inner sep=1.5pt, font=\sffamily\scriptsize]
      at (11.0,-1.8) {\textless\,3\,ms};
    \node[section] at (7.70,-1.42) {Computational Improvements};
    \node[secref] at (12.65,-1.32) {\ref{sec:computational_improvements}};
    \node[anchor=north west, align=left, inner sep=0pt,
          font=\sffamily\small\linespread{1.4}\selectfont]
      at (9.8,-1.65) {Fast\\[ -0.7ex ] fitting, adaptation, prediction};
    \node[img] at (10.7,-2.8)
      {\includegraphics[width=2.9\rkmpu,height=1.2\rkmpu,keepaspectratio, trim={155 0 155 0},clip,page=13]{bilder}};
    \node[rotate=-45, inner sep=1.5pt, font=\sffamily\scriptsize]
      at (10.8,-2.7) {\hypersetup{pdfborder={0 0 0}}\Cref{algo:ckmp:add_measurements}};
    \node[img] at (12.0,-2.8)
      {\includegraphics[width=2.9\rkmpu,height=1.2\rkmpu,keepaspectratio, trim={155 0 155 0},clip,page=14]{bilder}};
    \node[rotate=-45, inner sep=1.5pt, font=\sffamily\scriptsize]
      at (12.2,-2.6) {\hypersetup{pdfborder={0 0 0}}\Cref{algo:ckmp:remove_measurements}};
    \node[img] at (8.7,-2.5)
      {\includegraphics[width=2.9\rkmpu,height=1.8\rkmpu,keepaspectratio, trim={155 0 155 0},clip,page=12]{bilder}};

    \node[section] at (0.50,-3.92) {Time-based Policies};
    \node[img] at (2.23,-5.22)
      {\includegraphics[width=3.35\rkmpu,height=2.05\rkmpu,keepaspectratio,page=10, trim={60 20 20 40},clip]{bilder}};

    \node[boxtitle] at (5.70,-4.08) {Task Parameterization};
    \node[secref, anchor=north east] at (7.37,-3.86) {\ref{sec:kmp:task_parameterization}};
    \node[img] at (5.70,-5.1)
      {\includegraphics[width=3.50\rkmpu,height=1.50\rkmpu,keepaspectratio,page=11, trim={10 60 100 60},clip]{bilder}};

    \node[boxtitle] at (10.23,-4.10) {Adaptation};
    \node[secref, anchor=north east] at (12.72,-3.86) {\ref{sec:kmp:adaptation}};
    \node[img] at (10.28,-5.21)
      {\includegraphics[width=4.5\rkmpu,height=1.68\rkmpu,keepaspectratio,page=8, trim={0 0 0 170},clip]{bilder}};

    \draw[orangebox] (0.40,-1.13) rectangle (7.55,-3.54);
    \draw[orangebox] (0.40,-3.65) rectangle (12.83,-6.32);
  \end{tikzpicture}
  }
  \caption{Overview of our contributions. In \Cref{sec:kmp} we leverage Riemannian geometry to learn, \textcolor{dlrblue1}{adapt} and \textcolor{deeppastelpurple}{task-parameterize} demonstrated movement primitives including both aleatoric and epistemic uncertainties. \Cref{sec:matrix_kernels} further extends the framework to \textcolor{dlrgreen1}{\Acrshort{dof}-dependent} uncertainties. Finally, \Cref{sec:computational_improvements} improves the \textcolor{dlryellow1}{efficiency} of the required computations.\vspace{-1em}}
  \label{fig:intro:overview}
\end{figure}

Moreover, currently available kernel-based methods, using separable kernels, can only predict scalar-valued epistemic uncertainties as also seen in the decomposition in~\cite{knauer2025interactive}.
Recently, matrix-valued kernels have become available~\cite{alvarez2012kernels,micheli2013matrix} which allow for encoding correlations for same-sized inputs and outputs.
As a large step in the direction of such kernels, we leverage diagonal kernels to improve the learning of state-based \glspl{ds} (\Cref{sec:matrix_kernels}).

Finally, as in any non-parametric method, runtime in \glspl{kmp} is a function of the number of data points and in general slower than for parametric methods.
Especially adding or removing points is computationally expensive, requiring the inversion of a large matrix.
\citet{wang2026computationally} recently proposed a method for avoiding to recalculate the whole inverse.
Similar to \glspl{gp}~\cite{williams2006gaussian}, we leverage the Cholesky decomposition, significantly improving runtime over both vanilla \gls{kmp} and \cite{wang2026computationally} while keeping flexibility in adding and removing points (\Cref{sec:computational_improvements}).

In summary, the key contributions of our work are:
\begin{enumerate}
	\item A Riemannian formulation of a non-parametric approach to imitation learning \cite{huang2019kernelized} providing both aleatoric and epistemic heteroscedastic uncertainties (\Cref{sec:kmp:prediction});
	\item The Riemannian derivation of two adaptation schemes for different requirements based on via points \cite{huang2019kernelized} as well as on a soft null space projector \cite{silverio2023nonparametric} (\Cref{sec:kmp:adaptation});
	\item Task-parameterization of the method on different charts \cite{ti2023geometric} on $\mathrm{SE}(3)$ (\Cref{sec:kmp:task_parameterization});
	\item The integration of non-separable matrix-valued kernels in the approach to allow for the prediction of a matrix-valued epistemic uncertainty (\Cref{sec:matrix_kernels:diagonal});
	\item Computational improvements allowing for fast prediction and adaptation in several geometries commonly used in robotics problems (\Cref{sec:computational_improvements}).
\end{enumerate}

At the end of \Cref{sec:kmp,sec:matrix_kernels,sec:computational_improvements} we provide toy examples evaluating the key contributions in simplified scenarios.
Furthermore, we have implemented both time-based policies and their modification as well as state-based \glspl{ds} on a torque-controlled robot (\Cref{sec:evaluation}).
\Cref{sec:background} covers background.
We discuss results in \Cref{sec:limitations} and conclude in \Cref{sec:conclusion}.

\section{Background}
\label{sec:background}

\subsection{Riemannian Manifolds, Optimization and Statistics}
\label{sec:background:riemann}
\begin{figure}
\vspace{-0.5em}
\centering
\begin{subfigure}{0.32\columnwidth}
  \centering
  \includegraphics[width=\columnwidth,trim={120 65 120 65},clip]{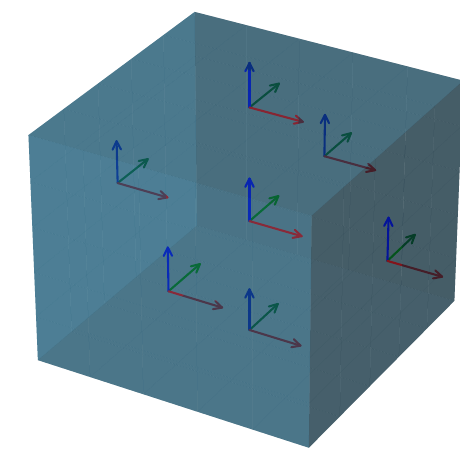}
  \subcaption{\tiny $\mathcal{M}_1$: $\mathbb{R}^3 \times \mathrm{SO}(3)$.}
\end{subfigure}
\hfill
\begin{subfigure}{0.32\columnwidth}
  \centering
  \includegraphics[width=\columnwidth,trim={120 65 120 65},clip]{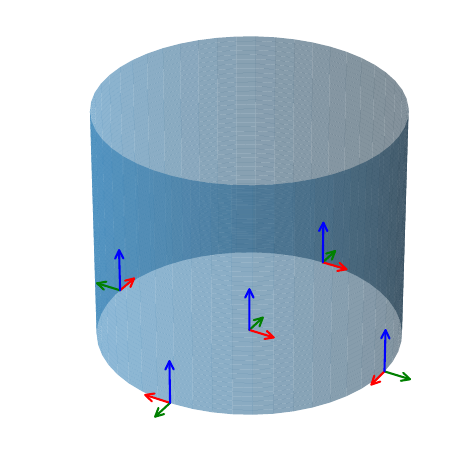}
  \subcaption{\tiny $\mathcal{M}_2$: $\mathcal{S}^1 \times \mathbb{R}^2 \times \mathrm{SO}(3)$.}
\end{subfigure}
\hfill
\begin{subfigure}{0.32\columnwidth}
  \centering
  \includegraphics[width=\columnwidth,trim={150 80 130 100},clip]{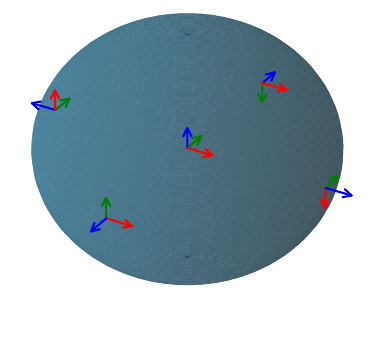}
  \subcaption{\tiny $\mathcal{M}_3$: $\mathcal{S}^2 \times \mathbb{R}^1 \times \mathrm{SO}(3)$.}
\end{subfigure}%
\caption{Manifolds used in this work, proposed by \cite{ti2023geometric}. Cylindrical ($\mathcal{M}_2$) or spherical ($\mathcal{M}_3$) coordinates allow expressing some tasks more data-efficient compared to Cartesian ($\mathcal{M}_1$) coordinates. Coordinate systems in each image depict the unit quaternion $(0, 0, 0, 1)^\top$ for different positions on the manifold.\vspace{-1em}}
\label{fig:background:manifolds}
\end{figure}
Many quantities in robotics, such as orientations in $\mathrm{SO}(3)$, cannot be modeled using Euclidean geometry.
Riemannian manifolds allow for a principled treatment of such non-Euclidean geometries \cite{calinon2020gaussians}.
Common to all smooth manifolds $\mathcal{M}$ is that the logarithm function $\bm{u}_{12} = \Log_{\bm{x}_1}^\mathcal{M} \left( \bm{x}_2 \right)$ maps a point $\bm{x}_2 \in \mathcal{M}$ onto the tangent space $\mathcal{T}_{\bm {x}_1}\mathcal{M}$ of $\mathcal{M}$ at $\bm{x}_1$, with the exponential function locally performing the inverse, calculating $\bm{x}_{2} = \Exp_{\bm{x}_1}^\mathcal{M} \left( \bm{u}_{12} \right)$.
\textit{Parallel transport} by the Levi-Civita connection moves tangent vectors between tangent spaces at different points on $\mathcal{M}$.
Optimization on manifolds can be performed using geodesic regression~\cite{fletcher2012geodesic} (multivariate case:~\cite{kim2014multvariate}), making use of \textit{parallel transport} to approximate a small update in the vicinity of the current linearization point.

We implement geometric priors by choosing different Riemannian metrics on $\mathrm{SE}(3)$, which we construct as \emph{pullback metrics} from charts of different \emph{representational manifolds} $\mathcal{M}_i$ as shown in \Cref{fig:background:manifolds}.
All manifolds $\mathcal{M}_i$ are six-dimensional product manifolds, and we construct maps $\mathcal{F}_i: \mathrm{SE}(3) \rightarrow \mathcal{M}_i$ that relate poses $\bm{x} \in \mathrm{SE}(3)$ to their representations $\bm{p} \in \mathcal{M}_i$, where the Jacobians $\bm{J}_{\mathcal{F}_i}$ as given in~\cite{muehlbauer2025unified} push tangent vectors between tangent spaces of $\mathcal{M}_i$. Further, we have charts $\varphi_i: \mathcal{U} \subset \mathcal{M}_i \rightarrow \varphi_i \left( \mathcal{U} \right) \subset \mathbb{R}^6$ for the representational manifolds, which can be constructed by stacking charts of the individual factor manifolds due to the product manifold structure of $\mathcal{M}_i$.
Using the concatenations $\psi_i = \varphi_i \circ \mathcal{F}_i$, we obtain charts $\psi_i: \mathcal{U}_i \subseteq \mathrm{SE}(3) \rightarrow \psi_i \left( \mathcal{U}_i \right) \subset \mathbb{R}^6$, with which we can pull the Euclidean metric of $\mathbb{R}^6$ back to $\mathrm{SE}(3)$ (first fundamental form). This induces different Riemannian metrics on $\mathrm{SE}(3)$ for each choice of \emph{representational manifold} $\mathcal{M}_i$, and an ordered basis for tangent spaces $\mathcal{T}_{\bm {x}}\mathcal{M}_i$.

The Riemannian Gaussian distribution \cite{zeestraten2017manifold,calinon2020gaussians} uses geodesic distances computed via the log map based on this metric
\begin{gather}
    \mathcal{N}\left(\bm{x}\vert\bm{\mu},\bm{\Sigma}\right) = \frac{1}{\sqrt{\left(2\pi\right)^D\mathrm{det}(\bm{\Sigma})}}e^{-\frac{1}{2}\mathrm{Log}_{\bm{\mu}}^{\mathcal{M}_i}\!\left(\bm{x}\right)^\top\bm{\Sigma}^{-1}\mathrm{Log}_{\bm{\mu}}^{\mathcal{M}_i}\!\left(\bm{x}\right)},
    \raisetag{12pt}\label{eq:background:riemannian_gaussian}
\end{gather}
with mean $\bm{\mu}\in\mathcal{M}_i$, covariance $\bm{\Sigma} \in \mathcal{T}_{\bm{\mu}} \mathcal{M}_i \otimes \mathcal{T}_{\bm{\mu}} \mathcal{M}_i$ and the dimension $D$ of $\mathcal{M}_i$.
Using the metric induced by the charts defined above, we get different distances and thus different Gaussian shapes depending on the chosen manifold.
As a limitation of the Riemannian Gaussian, \eqref{eq:background:riemannian_gaussian} is only well-defined for covariances small enough that $\mathrm{Exp}$ is a diffeomorphism.

\subsection{Probabilistic Learning from Demonstration}
One key goal in probabilistic movement modeling, besides an accurate reconstruction, is to provide an estimate of uncertainty.
\glspl{gmm}~\cite{calinon2015tutorial,calinon2020gaussians} modeling \textit{aleatoric uncertainty} encode the joint distribution between input $\bm{s}$ and output $\bm{\xi}$ with $M$ Gaussians and mixing coefficients $\pi_m$ ($\sum_{m=1}^M \pi_m = 1$),
\begin{equation}
		\label{eq:background:probabilistic:joint_gmm_encoding}
	\begin{bmatrix} \bm{s} \\ \bm{\xi} \end{bmatrix} \sim \sum_{m=1}^{M} \pi_m \mathcal{N} \left(\left.\begin{bmatrix} \bm{s} \\ \bm{\xi} \end{bmatrix}\right| \bm{\mu}_m,\bm{\Sigma}_m\right).
\end{equation}
The conditional distribution of $\bm{\xi}$ given $\bm{s}$ evaluates to
\begin{equation}
	p(\bm{\xi} \vert \bm{s}) = \sum_{m=1}^{M} \pi_m \mathcal{N} (\bm{\xi} \vert \bm{\mu}_{m|\bm{s}},\bm{\Sigma}_{m|\bm{s}})
	\label{eq:background:probabilistic:gmm_multimodal}
\end{equation}
using \gls{gmr}.
The extension of \gls{gmm} / \gls{gmr} to Riemannian manifolds is given by \cite{zeestraten2017manifold,calinon2020gaussians}.

\glspl{gp} \cite{williams2006gaussian} compute posterior mean and covariance of a function $\bm{\xi}(\bm{s}) = f(\bm{s}) + \epsilon$, modeling the \textit{epistemic uncertainty}; an extension to Riemannian manifolds is given in~\cite{mallasto2018wrapped}.
Based on a probabilistic reference $\mathcal{D} = \{\bm{s}_n, \bm{\mu}_n,\bm{\Sigma}_n\}_{n=1}^N$ obtained through \gls{gmm} / \gls{gmr}, \glspl{kmp} \cite{huang2019kernelized} model both \textit{aleatoric} and \textit{epistemic uncertainty}.
For a test point $\bm{s}^*$, we obtain
\begin{align}
	\mathrm{E} \left[ \bm{\xi}^*(\bm{s}) \right] &= \bm{K}(\bm{s}^*, \bm{s}) \left( \bm{K}(\bm{s}, \bm{s}) + \lambda \bm{\Sigma} \right)^{-1} \bm{\mu}\label{eq:background:kmp_pred_1}\\
	\mathrm{Cov} \left[ \bm{\xi}^*(\bm{s}) \right] &= \alpha \left. \Bigl( \bm{K}(\bm{s}^*, \bm{s}^*) \right. \label{eq:background:kmp_pred_2}\\
	 &~~~- \left. \bm{K}(\bm{s}^*, \bm{s}) \left( \bm{K}(\bm{s}, \bm{s}) + \lambda_c \bm{\Sigma} \right)^{-1} \bm{K}(\bm{s}, \bm{s}^*) \right), \nonumber
\end{align}
where $\bm{K}$ is composed of kernel matrices, ${\bm{\mu}=[\bm{\mu}^\top_1,\ldots,\bm{\mu}^\top_N]^\top}$ and ${\bm{\Sigma}=\mathrm{blockdiag}\left(\bm{\Sigma}_1,\ldots,\bm{\Sigma}_N\right)}$ are set from the probabilistic reference.
$\lambda$, $\lambda_c$ and $\alpha$ are regularization and scaling hyperparameters.
By predicting full, input-dependent covariance matrices \eqref{eq:background:kmp_pred_2}, \glspl{kmp} are \textit{heteroscedastic} models.
Please see \cite{huang2019kernelized} for a detailed derivation. Note the connection to \glspl{gp} \cite{williams2006gaussian}, where we have $\lambda_c=\lambda$ and $\bm{\Sigma}=\bm{I}$ (\textit{homoscedasticity}), $\alpha=1$ and $\bm{\mu}$ are directly the observations $\bm{\xi}(\bm{s})$.

\section{Riemannian Kernelized Movement Primitives}
\label{sec:kmp}
We model the distribution of a random variable $\bm{\xi}(\bm{s})$
\begin{equation}
	p(\bm{\xi}(\bm{s}) | \bm{s}) = \mathcal{N}(\bm{\xi}(\bm{s}) | \bm{\mu}, \bm{\Sigma}),
\end{equation}
where $\bm{s}$ and $\bm{\xi}(\bm{s})$ are Euclidean or manifold-valued (\Cref{fig:background:manifolds}) and $\bm{\Sigma}$ contains both aleatoric and epistemic uncertainty.
In the following, we provide a derivation based on encoding a probabilistic reference $\mathcal{D} = \{ \bm{s}_n, \bm{\mu}_n, \bm{\Sigma}_n \}_{n=1}^N$ with $N$ entries of observations $\bm{\mu}_n \in \mathcal{M}$ with known covariance $\bm{\Sigma}_n \in \mathcal{T}_{\bm{\mu}_n} \mathcal{M} \otimes \mathcal{T}_{\bm{\mu}_n} \mathcal{M}$ for predicting new outputs (\ref{sec:kmp:prediction}), the adaptation to via points (\ref{sec:kmp:adaptation}), its task parameterization (\ref{sec:kmp:task_parameterization}) as well as an evaluation in simplified yet representative scenarios (\ref{sec:kmp:experiments}).

\subsection{Movement Primitive Prediction}
\label{sec:kmp:prediction}
As in \cite{huang2019kernelized,huang2021toward,silverio2023nonparametric} we assume a parametric function
\begin{gather}
	\bm{\xi}(\bm{s}) = \Exp_{\bm{e}}^{\mathcal{M}} \Biggl( {\underbrace{\begin{bmatrix}
		\bm{\phi}(\bm{s}) & \dots & \bm{0}\\
		\vdots & \ddots & \vdots\\
		\bm{0} & \dots & \bm{\phi}(\bm{s})
	\end{bmatrix}}_{\bm{\Phi}(\bm{s})}}^\top \bm{w} \Biggr) = \Exp_{\bm{e}} \left( \bm{\Phi}(\bm{s})^\top \bm{w} \right)\label{eq:weight_space:prediction}
	\raisetag{12pt}
\end{gather}
where $\bm{\phi}(\bm{s}) \in \mathbb{R}^B$ is a $B$-dimensional vector of basis functions depending on $\bm{s}$. $\bm{w} \in \mathbb{R}^{BD}$ is a vector of parametric weights, with the tangent dimension $D$ of $\bm{\xi}(\bm{s}) \in \mathcal{M}$ and $\bm{e} \in \mathcal{M}$ the process mean function value.
Assuming normally distributed weights $\bm{w}$ with $p(\bm{w}) = \mathcal{N}(\bm{\mu}_w, \bm{\Sigma}_w)$, the distribution in \eqref{eq:weight_space:prediction} is also normally distributed~\cite{huang2019kernelized}.
As in vanilla \gls{kmp}, we seek to minimize the \Acrlong{kl} divergence between reference and predictive distributions.
To this end, we maximize the posterior probability by solving the regularized weighted least squares problem with residual vectors calculated in $\mathcal{T}_{\bm{\mu}_n}\mathcal{M}$, while $\bm{\mu}_w$ is expressed in $\mathcal{T}_{\bm{e}}\mathcal{M}$,
\begin{align}
	J(\bm{\mu}_w) &= \sum_{n=1}^N \left( \Log_{\bm{\mu}_n} \left( \Exp_{\bm{e}} \left( \bm{\Phi}(\bm{s}_n)^\top \bm{\mu}_w \right) \right)^\top \bm{\Sigma}_n^{-1} \right.\label{eq:weight_space:costfcn_orig}\\
	& ~~~~~~~~~~ \left. \Log_{\bm{\mu}_n} \left( \Exp_{\bm{e}} \left( \bm{\Phi}(\bm{s}_n)^\top \bm{\mu}_w \right) \right) \right) + \lambda \bm{\mu}_w^\top \bm{\mu}_w.\nonumber
\end{align}
Taking inspiration from \cite{kim2014multvariate}, we approximate this nonlinear cost function by a quadratic and kernelizable problem
\begin{align}
	J(\bm{\mu}_w) & \approx \sum_{n=1}^N \left( \left( \Gamma_{\bm{\mu}_n} \left( \bm{\Phi}(\bm{s}_n)^\top \bm{\mu}_w \right) + \Log_{\bm{\mu}_n} \bm{e} \right)^\top \bm{\Sigma}_n^{-1} \right.\label{eq:weight_space:costfcn}\\
	& ~~~~~~~~~~ \left. \left( \Gamma_{\bm{\mu}_n} \left( \bm{\Phi}(\bm{s}_n)^\top \bm{\mu}_w \right) + \Log_{\bm{\mu}_n} \bm{e} \right) \right) + \lambda \bm{\mu}_w^\top \bm{\mu}_w,\nonumber
\end{align}
where $\Gamma_{\bm{\mu}_n} \left( \cdot \right)$ denotes the parallel transport $\Gamma_{\bm{e} \rightarrow \bm{\mu}_n} \left( \cdot \right)$ from the mean function value $\bm{e}$ to $\bm{\mu}_n$.
Note that this cost function is never evaluated explicitly as the basis functions project into an infinite-dimensional feature space.
Comparing the kernelized result with the function-space derivation of Riemannian \glspl{kmp} in \Cref{sec:appendix:prediction_conditioning} shows that indeed, no precision is lost by this approximation.
Unlike the weight-space derivation in \cite{huang2021toward}, we do not linearize the problem in a single tangent space $\mathcal{T}_{\bm{x}_0}\mathcal{M}$ at an arbitrary $\bm{x}_0$.
Instead, we can utilize Riemannian Gaussians with manifold-valued mean $\bm{\mu}_n$ and covariance expressed in $\mathcal{T}_{\bm{\mu}_n} \mathcal{M} \otimes \mathcal{T}_{\bm{\mu}_n} \mathcal{M}$ as reference, e.g., obtained from a Riemannian Gaussian Mixture Regression \cite{zeestraten2017manifold}.
We furthermore do not restrict the solution to quaternions, but remain general to Riemannian manifolds where the parallel transport $\Gamma_{\bm{e} \rightarrow \bm{\mu}_n} \left( \cdot \right)$ can be expressed using left-multiplication of a rotation matrix $\bm{R}_{\bm{e} \rightarrow \bm{\mu}_n}$.
Differentiating the objective \eqref{eq:weight_space:costfcn} and equating to zero thus results in
\begin{align}
	\left( \left( \bm{R}_{\bm{e} \rightarrow \bm{\mu}} \bm{\Phi}(\bm{s})^\top \right)^\top \bm{\Sigma}^{-1} \left( \bm{R}_{\bm{e} \rightarrow \bm{\mu}} \bm{\Phi}(\bm{s})^\top \right) + \lambda \bm{I} \right) \bm{\mu}_w^* =\nonumber\\
	-\left( \bm{R}_{\bm{e} \rightarrow \bm{\mu}} \bm{\Phi}(\bm{s})^\top \right)^\top \bm{\Sigma}^{-1} \Log_{\bm{\mu}}(\bm{e}),\label{eq:weight_space_derivative}
\end{align}
where stacked matrices from the probabilistic reference
\begin{gather}
	\bm{\Phi} = \left[ \bm{\Phi}(\bm{s}_1) \hdots \bm{\Phi}(\bm{s}_N) \right], \quad \bm{\mu} = \left[ \bm{\mu}_1 \hdots \bm{\mu}_N \right]^\top,\nonumber\\
	\bm{\Sigma} = \mathrm{blockdiag}(\bm{\Sigma}_1 \hdots \bm{\Sigma}_N)
\end{gather}
have been used.
Using the matrix identity \cite{bishop2006pattern}
\begin{equation}
	\left( \bm{B}^\top \bm{R}^{-1} \bm{B} + \bm{P}^{-1} \right)^{-1} \bm{B}^\top \bm{R}^{-1} = \bm{P} \bm{B}^\top (\bm{B} \bm{P} \bm{B}^\top + \bm{R})^{-1},\nonumber
\end{equation}
and setting $\bm{\Phi}_\parallel(\bm{s}_n) = \bm{\Phi}(\bm{s}_n) \bm{R}_{\bm{e} \rightarrow \bm{\mu}_n}^\top$ we arrive at the solution
\begin{equation}
	\bm{\mu}_w^* = -\bm{\Phi}_\parallel(\bm{s}) \left( \bm{\Phi}_\parallel(\bm{s})^\top \bm{\Phi}_\parallel(\bm{s}) + \lambda \bm{\Sigma} \right)^{-1} \Log_{\bm{\mu}}(\bm{e}),\label{eq:weight_space:solution}
\end{equation}
Substituting \eqref{eq:weight_space:solution} in \eqref{eq:weight_space:prediction} and applying the \textit{kernel trick} \cite{bishop2006pattern},
\begin{align}
	\mathbb{E} \left[ \bm{\xi}(\bm{s}^*) \right] &= \Exp_{\bm{e}} \left( \bm{\Phi}_\parallel(\bm{s}^*)^\top \bm{\mu}_w^* \right) \nonumber\\
	&= \Exp_{\bm{e}} \left( -\bm{k}_\parallel^* \left( \bm{K}_\parallel + \lambda \bm{\Sigma} \right)^{-1} \Log_{\bm{\mu}}(\bm{e}) \right) \label{eq:kmp_mean_prediction}
\end{align}
follows for the expectation which is the same result as in \Cref{sec:appendix:prediction_conditioning}, where the \gls{kmp} is derived as a Gaussian conditioning.
Note that on the manifold $\mathbb{R}^n$, \eqref{eq:kmp_mean_prediction} evaluates to the same result as the Euclidean \gls{kmp} prediction \eqref{eq:background:kmp_pred_1} - our Riemannian derivation thus contains the Euclidean variant as a special case.
The covariance evaluates to (\Cref{sec:appendix:prediction_conditioning})
\begin{equation}
	\bm{\Sigma}(\bm{s}^*) = \bm{k}_{\parallel}(\bm{s}^*, \bm{s}^*) - \bm{k}_{\parallel}^* \left(\bm{K}_{\parallel} + \lambda\bm{\Sigma}\right)^{-1} \bm{k}_{\parallel}^{*\top}.
\label{eq:kmp_covariance_prediction}
\end{equation}

Calculating the mean via \eqref{eq:kmp_mean_prediction} suffers from imprecision for certain manifolds when predicting a new point close to an antipodal point -- e.g., on $\mathrm{SO}(3)$, tangent entries of $-\pi$ and $\pi$ denote the same orientation change.
We can often avoid these cases by recentering around the Fréchet mean of the data.

When predicting the orientation on $\mathrm{SO}(3)$, we can also make use of the double coverage of $\mathrm{SO}(3)$ by $\mathcal{S}^3$.
Leveraging the whole range of $\left[-2\pi, 2\pi\right[$ for values in $\mathcal{T} \mathcal{S}^3$ instead of just $\left[-\pi, \pi\right[$ for $\mathcal{T} \mathrm{SO}(3)$, we can ``push'' the discontinuity between $-\pi$ and $\pi$ away from query input.
If the trajectory involves orientation changes of less than $\SI{360}{\degree}$, one updated instance of $\Log_{\bm{\mu}}(\bm{e})$ is sufficient.
For larger orientation changes, we can create multiple instances where we always select the instance with the discontinuity furthest from the query input.

\subsection{Movement Primitive Adaptation}
\label{sec:kmp:adaptation}
Two main approaches to adapt \glspl{kmp} to novel situations exist, suiting different types of use cases.

\paragraph{Via-point based adaptation}
As in the original \gls{kmp} \cite{huang2019kernelized}, adaptation can be performed by extending the reference $\mathcal{D}$ by a \textit{desired database} $\overline{\mathcal{D}} =\{ \overline{\bm{s}}_p, \overline{\bm{\mu}}_p, \overline{\bm{\Sigma}}_p \}_{p=1}^P$.
Low covariances $\overline{\bm{\Sigma}}_p$ result in a modification close to $\overline{\bm{\mu}}_p$, larger covariances in a prediction close to the original one.
Augmenting the \gls{kmp} with $\overline{\mathcal{D}}$ requires the extension of $\bm{K} + \lambda \bm{\Sigma}$ and the costly recomputation of its inverse.
In \Cref{sec:computational_improvements}, we provide computationally faster alternatives for calculating the update, allowing for a quick update, e.g., through human correction.

\paragraph{Modulation with uncertainty-aware deformations}
Soft null space adaptations \cite{silverio2023nonparametric} guide the adaptation by the previously learned covariance; they have been used to constrain reinforcement learning \cite{padalkar2025towards} or coarse human modifications \cite{quere2024probabilistic}.
The goal of the adaptation \cite{silverio2023nonparametric} is to keep weights $\bm{\mu}_w$ in \eqref{eq:weight_space:costfcn} close to desired weights $\hat{\bm{w}}$, which can be achieved by adding the penalty term $\beta (\bm{\mu}_w - \hat{\bm{w}})^\top (\bm{\mu}_w - \hat{\bm{w}})$ to \eqref{eq:weight_space:costfcn}.
For a desired modification $\hat{\bm{\xi}} \in \mathcal{M}$, similar to the Euclidean derivations in \cite{silverio2023nonparametric}, we can calculate a new expected value $\bm{\xi}'(\bm{s}^*)$ (detailed in \Cref{sec:appendix:soft_nullspace}), extending \eqref{eq:weight_space:prediction}
\begin{align}
	\mathbb{E} \left[ \bm{\xi}'(\bm{s}^*) \right] &= \mathrm{Exp}_{\bm{e}}\left(-\bm{k}_\parallel^* \bm{\Psi}_\parallel \mathrm{Log}_{\bm{\mu}}(\bm{e}) +\left[ \hat{\bm{k}}_\parallel^* - \bm{k}_\parallel^* \bm{\Psi}_\parallel \hat{\bm{K}}_\parallel \right]\right.\nonumber\\
	& ~~~ \left. \left(\underline{\bm{K}}_\parallel\right)^{-1} \Gamma_{\bm{e} \rightarrow \bm{\xi}(\bm{s}^*)} \left( \Log_{\bm{\xi}(\bm{s}^*)}(\hat{\bm{\xi}})\right)\right)\label{eq:nullspace:prediction}
\end{align}
where $\bm{\Psi}_\parallel = \left( \bm{K}_\parallel + \lambda \bm{\Sigma} \right)^{-1}$, $\hat{\bm{K}}_\parallel = \bm{\Phi}_\parallel^\top \hat{\bm{\Phi}}_\parallel$ and $\underline{\bm{K}}_\parallel = \hat{\bm{\Phi}}_\parallel^\top \hat{\bm{\Phi}}_\parallel$ and $\bm{\xi}(\bm{s}^*)$ is the prediction before applying the modification.
Owing to the uncertainty-aware deformation, $\Log_{\bm{\xi}(\bm{s}^*)}(\hat{\bm{\xi}})$ is in practice scaled to achieve the desired deformation as in the Euclidean case \cite{silverio2023nonparametric}.
In learning approaches such as \cite{padalkar2025towards}, we can use $\Log_{\bm{\xi}(\bm{s}^*)}(\hat{\bm{\xi}})$ as \textit{action space} for a covariance-guided exploration.

\begin{figure}
	\vspace{-1em}
	\centering
	\begin{subfigure}{0.49\columnwidth}
		\annotategraphicsmulti{
			\includegraphics[width=\columnwidth,trim={105 30 70 27},clip]{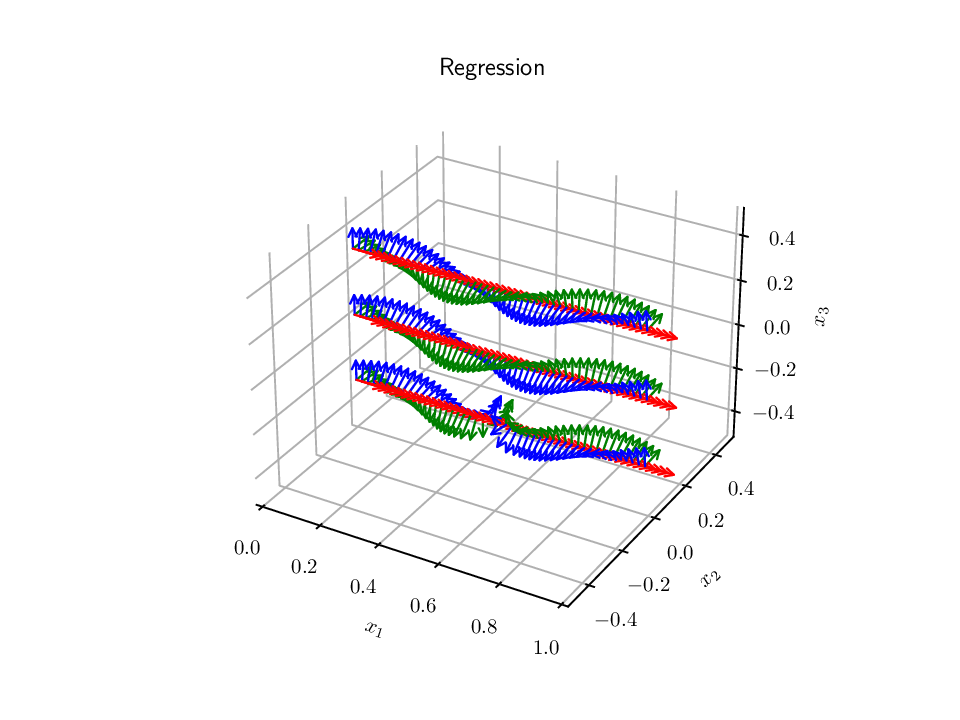}
		}{
        		\node at (0.15,0.72) {1};
        		\node at (0.15,0.59) {2};
        		\node at (0.15,0.46) {3};
    }
	\end{subfigure}
	\begin{subfigure}{0.49\columnwidth}
		\annotategraphicsmulti{
			\includegraphics[width=\columnwidth,trim={105 30 70 27},clip]{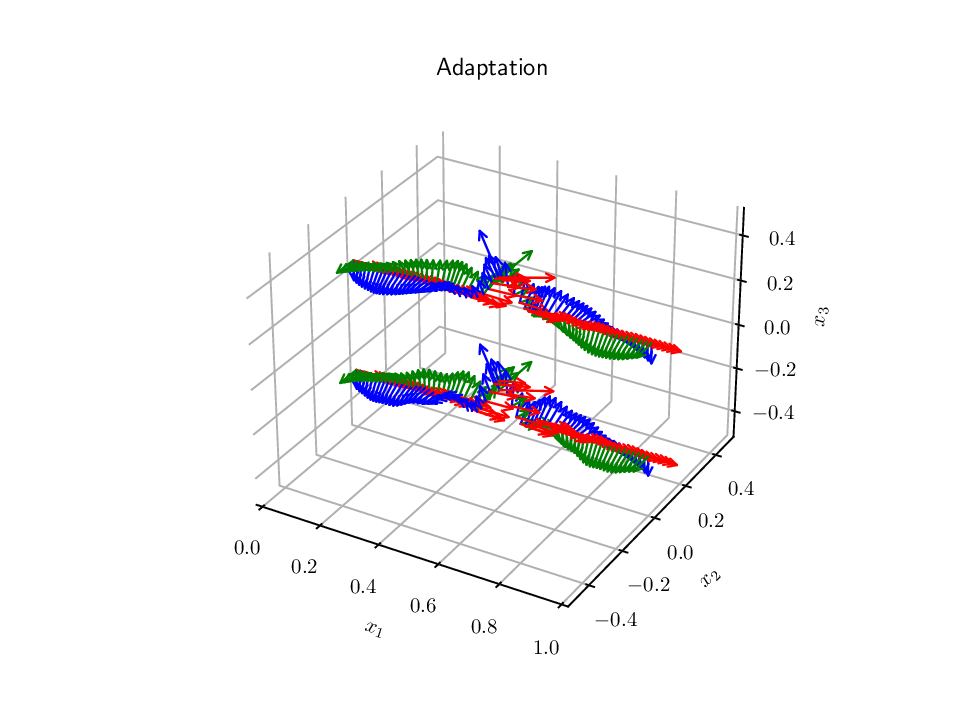}
		}{
        		\node at (0.15,0.68) {V};
        		\node at (0.15,0.48) {N};
    }
	\end{subfigure}
	\caption{\label{fig:kmp:line} \textit{Left}: Pose data \protect\lref{1} reproduced by a Riemannian \gls{kmp}~(\ref{sec:kmp:prediction}) \protect\lref{2} with correct rotations. In contrast, the \gls{sts} \gls{kmp}~\cite{huang2021toward} \protect\lref{3} suffers from incorrect rotations. \textit{Right}: Adding a via point \protect\lref{V} \SI{0.1}{\metre} above the trajectory with a relative rotation of \SI{-0.35}{\radian} about the $y$ axis and a null space action \protect\lref{N}. Via point and nullspace target are visualized by a larger coordinate system.\vspace{-1em}}
\end{figure}
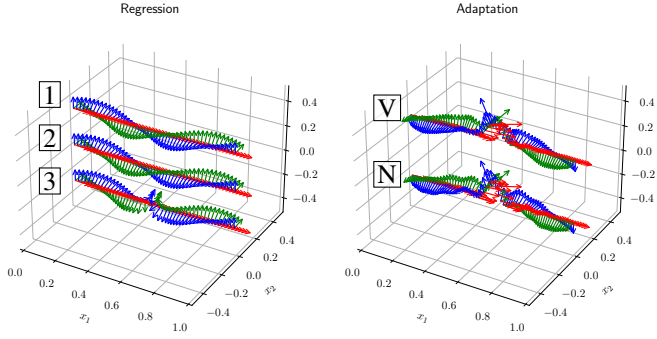

\subsection{Task Parameterization}
\label{sec:kmp:task_parameterization}
Using task parameterization \cite{calinon2015tutorial}, we can learn movement models locally, with respect to different coordinate origins $\bm{x}_i \in \mathrm{SE}(3)$.
Defining new values of these origins at execution time, e.g., due to new object positions/orientations, facilitates skill generalization.
Such $\mathrm{SE}(3)$ transformations equate to a rotation $\bm{A} \in \mathbb{R}^{D \times D}$ of the tangent space and a translation $\bm{b} \in \mathcal{M}$~\cite{zeestraten2017manifold} with $\bm{\mu} \in \mathcal{M}$ and $\bm{\Sigma} \in \mathcal{T}_{\bm{\mu}} \mathcal{M} \otimes \mathcal{T}_{\bm{\mu}} \mathcal{M}$
\begin{align}
	\bm{\mu}_{\bm{A}, \bm{b}} &= \Exp_{\bm{b}} \left( \bm{A} \Log_{\bm{e}} \left( \bm{\mu} \right) \right),\label{eq:tp_mean}\\
	\bm{\Sigma}_{\bm{A}, \bm{b}} &= \bm{A} \bm{\Sigma} \bm{A}^\top.\label{eq:tp_cov}
\end{align}
Note that unlike in \cite{zeestraten2017manifold}, the required parallel transport evaluates to identity on $\mathcal{M}_1$,\footnote{On $\mathcal{S}^2$, parallel transport is required due to the hairy ball theorem.\label{footnote:hairy_ball}} as tangent spaces are identified trivially on $\mathbb{R}^3$ respectively via the quaternion product on $\mathcal{S}^3$.
Local models on $\mathcal{M}_1$ can be rotated by setting $\bm{A} = \mathrm{blockdiag} \left( \bm{R}_{\bm{e}\bm{b}}, \bm{I} \right)$ corresponding to the rotation of position \glspl{dof} in a $\mathrm{SE}(3)$ transformation and translated by the new mean $\bm{b}$.
After transforming all Gaussian predictions into a common coordinate system by applying \eqref{eq:tp_mean} -- \eqref{eq:tp_cov}, the Riemannian Gaussian product \cite{zeestraten2017manifold} allows for fusing the prediction.

On $\mathcal{M}_2$ and $\mathcal{M}_3$, translating the origin would invalidate covariances, as angular \glspl{dof}, radius and identity orientation depend on the coordinate origin.
We can however still exploit the properties of one model expressed in $\mathcal{M}_2$ or $\mathcal{M}_3$, fusing it with one or multiple models in $\mathcal{M}_1$.
To this end, we use the coordinate origin of the model on $\mathcal{M}_2$ or $\mathcal{M}_3$ as global origin and move the other model outputs using \eqref{eq:tp_mean} -- \eqref{eq:tp_cov} to this new origin, still expressing them on $\mathcal{M}_1$.
Subsequently, using the coordinate transformation formulas and Jacobians given in \cite{muehlbauer2025unified}, we transform means $\bm{\mu}_{\mathcal{M}_1}$ and covariances $\bm{\Sigma}_{\mathcal{M}_1}$ to $\bm{\mu}_{\mathcal{M}_{2,3}}$ and $\bm{\Sigma}_{\mathcal{M}_{2,3}}$.
Finally, a Gaussian product on $\mathcal{M}_2$ or $\mathcal{M}_3$ fuses the result.
This result can then be used to subsequently leverage properties of different $\mathcal{M}$, e.g., manifold-aware variable stiffness as presented in \cite{muehlbauer2025unified}.

\subsection{Evaluation in Toy Examples}
\label{sec:kmp:experiments}
\begin{figure}
\vspace{-1em}
\centering
\begin{subfigure}{0.24\textwidth}
  \centering
  \includegraphics[width=\columnwidth,trim={120 72 90 45},clip]{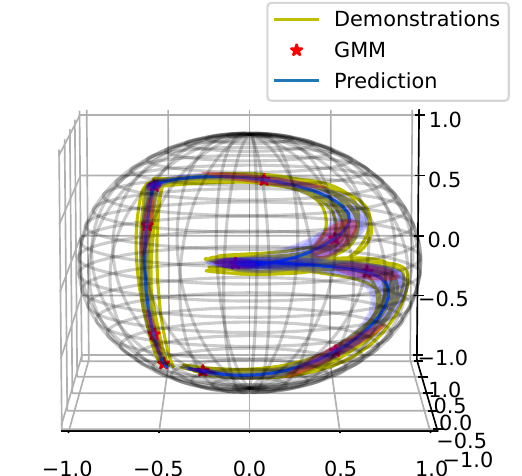}
  \subcaption{\Acrlong{gmr}.}
\end{subfigure}
\hfill
\begin{subfigure}{0.24\textwidth}
  \centering
  \includegraphics[width=\columnwidth,trim={120 72 90 45},clip]{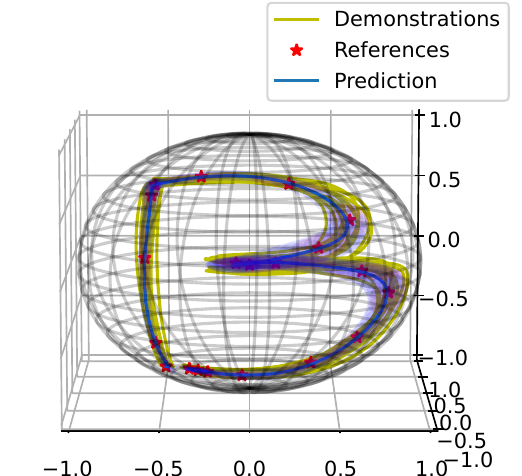}
  \subcaption{\Acrlong{kmp}.}
\end{subfigure}%
\caption{Plots for the letter ``B'' of the handwriting dataset \cite{silverio2023nonparametric} projected to $\mathcal{S}^2$. We fit the \gls{gmm} using $10$ Gaussians and use $20$ reference points for \gls{kmp}; all these Gaussians are depicted with their mean and covariance ellipsoid.\vspace{-0.5em}}
\label{fig:eval:plots_s2}
\end{figure}
\begin{table}
\scriptsize
\caption{\label{tab:kmp:line}\centering \scshape Differences between the prediction of different methods and ground truth for the spiral trajectory (\Cref{fig:kmp:line}).}
\centering
\begin{tabular}{@{}c|cccc@{}}
Method & Position Difference (mm) & Rotation Difference (mrad)\\
\hline
\gls{gmr} (\gls{sts}) & $2.47 \pm 1.02$ & $3139.36 \pm 11.47$\\
\gls{kmp} (\gls{sts}) \cite{huang2021toward} & $2.47 \pm 1.03$ & $269.36 \pm 646.48$\\
Riemannian \gls{gmr} \cite{zeestraten2017manifold} & $2.47 \pm 1.09$ & $15.27 \pm 8.81$\\
Riemannian \gls{kmp} (ours) & $\bm{2.46 \pm 1.12}$ & $\bm{15.15 \pm 9.05}$\\
Nadaraya-Watson \cite{lopez2023non} & $3.05 \pm 1.22$ & $15.29 \pm 9.15$
\end{tabular}
\vspace{-1em}
\end{table}
To evaluate our Riemannian \gls{kmp} prediction (\Cref{sec:kmp:prediction}), we compare it to \gls{sts} baselines~\cite{huang2021toward} and a Riemannian \gls{gmr}~\cite{zeestraten2017manifold}.\footnote{While including Riemannian \gls{gmr} \cite{zeestraten2017manifold} in \Cref{tab:kmp:line}, we do not seek to outperform it in general as the probabilistic reference is extracted from \gls{gmr}, but to achieve comparable performance in regions of \textit{known} data.}
We encode a circular trajectory with $\bm{s} \in [0, 1]$ and an orientation change of $2 \pi$ depicted as the uppermost line in \Cref{fig:kmp:line}.
We sample $1000$ poses of this trajectory modeled from $100$ Gaussians centered at the nominal trajectory.
After fitting a \gls{gmm} with $5$ Gaussians, we extract a probabilistic reference composed of $50$ poses using \gls{gmr} on equidistant time inputs $\bm{s}^* \in [0, 1]$ to fit the \gls{kmp}.
We continue using this approach of extracting a probabilistic reference throughout the experimental evaluation.
For the \gls{sts} approach, we prepare data points using $\bm{\xi}\left(\bm{s}\right)_\mathrm{sts} = \Log_{\bm{e}} \left( \bm{\xi}\left(\bm{s}\right)_{\mathcal{M}_1} \right)$ and transform the prediction using $\bm{\xi}\left(\bm{s}^*\right)_{\mathcal{M}_1} = \Exp_{\bm{e}} \left( \bm{\xi}\left(\bm{s}^*\right)_\mathrm{sts} \right)$.
For our Riemannian approach, we employ \eqref{eq:kmp_mean_prediction}.
Both use a \gls{rbf} kernel with $l=0.1$.
The results are visualized in \Cref{fig:kmp:line} -- while our approach is able to predict all orientations correctly, the \gls{sts} approach fails completely for some orientations.
As additional baseline, we use a Riemannian Nadaraya-Watson regression \cite{muehlbauer2026geometric} on all sampled poses and a \gls{rbf} kernel with $l=0.007$.
We omit plotting the Nadaraya-Watson result as it is visually indistinguishable from the \gls{kmp} result.
\Cref{tab:kmp:line} summarizes positional and rotational error for all approaches averaged over $200$ query points - while all models achieve a good accuracy for position prediction, the orientation prediction of the \gls{sts} \gls{kmp}~\cite{huang2021toward} suffers from discontinuities of the log map.
We achieve a slightly better accuracy than the Nadaraya-Watson baseline, even though it has access to all data points.
Note that this baseline cannot model epistemic uncertainty which is a key requirement e.g., in state-based policies; furthermore it has an increased runtime with many sample points as seen in \Cref{sec:computational_improvements:adding:experiments}.

Figure~\ref{fig:kmp:line} furthermore shows the adaptation using a via point above the original trajectory, including an orientation change, and a similar null space action (\Cref{sec:kmp:adaptation}).

Another evaluation of our proposed Riemannian \gls{kmp} on $\mathcal{S}^2$ is depicted in \Cref{fig:eval:plots_s2}, showcasing that the method works on different Riemannian manifolds and is not constrained to manifolds exhibiting a Lie group structure.

\begin{nodiff}

\begin{table}
\scriptsize
\caption{\label{tab:matrix_kernels:length_scales}\centering \scshape Maximum length scales for selected kernels.}
\centering
\begin{tabular}{@{}c|cccccc@{}}
Kernel & $\mathcal{S}^1$ & $\mathcal{S}^1 \times \mathbb{R}^2$ & $\mathcal{S}^2$ & $\mathrm{SO}(3)$ & $\mathbb{R}^3 \times \mathrm{SO}(3)$\\
\hline
\gls{rbf} & $0.01$ & $0.65$ & $0.27$ & $0.63$ & $1.1$\\
Matérn ($\nu = \frac{3}{2}$) & $0.40$ & $0.99$ & $0.77$ & $0.97$ & $1.46$\\
Matérn ($\nu = \frac{5}{2}$) & $0.07$ & $0.83$ & $0.58$ & $0.8$ & $1.33$\\
\end{tabular}
\vspace{-1em}
\end{table}

\section{Matrix-valued Kernels}
\label{sec:matrix_kernels}
Commonly employed kernels for vector-valued output, such as separable kernels (\Cref{sec:matrix_kernels:separable}) used in \textit{vanilla} \gls{kmp} \cite{huang2019kernelized}, do not measure per-\gls{dof} uncertainties.
Instead, the output of a kernel with scalar output is assigned to all \glspl{dof}.
To address this shortcoming, matrix-valued kernels measuring per-\gls{dof} uncertainties compatible with our proposed Riemannian formulation are required.
In \Cref{sec:matrix_kernels:diagonal}, we introduce a novel diagonal kernel computing this required covariance.
We show the resulting differences on toy examples in \Cref{sec:matrix_kernels:experiments}.

\subsection{Separable Kernels}
\label{sec:matrix_kernels:separable}
A common choice for vector-valued output is the product of a kernel with scalar output and the identity matrix \cite{huang2019kernelized}
\begin{equation}
	\bm{k}(\bm{s}, \bm{s}') = k(\bm{s}, \bm{s}') \bm{I}.
\end{equation}
This type of kernel is called \textit{separable} as the interactions between outputs are modeled without reference to the input space~\cite{alvarez2012kernels}.
As such, these kernels can only output a scalar-valued epistemic uncertainty which is the same for all \glspl{dof}.

The resulting kernel matrix has only entries on its diagonal which are all identical.
Parallel transporting results in
\begin{equation}
	\bm{k}_\parallel(\bm{s}, \bm{s}') = \bm{R}_{\bm{e}\rightarrow\bm{\mu}} k(\bm{s}, \bm{s}') \bm{I} \bm{R}_{\bm{e}\rightarrow\bm{\mu}}^\top = k(\bm{s}, \bm{s}') \cdot \bm{I}
\end{equation}
and thus does not lead to changes, irrespective of the chosen tangent space.
Using the $6 \times 6$ matrix $\bm{I}_6$, this formulation can, e.g., be used to predict full pose outputs.

To use manifold-valued inputs in the \gls{rbf} and Matérn kernels, we use the scaled Riemannian distance $\sfrac{d^\mathcal{M}(\bm{s}, \bm{s}')}{l} = \sfrac{\left|\left| \Log^{\mathcal{M}}_{\bm{s}} \left( \bm{s}' \right) \right|\right|_2}{l} = r$ to compute
\begin{align}
	k_\mathrm{RBF}^\mathcal{M}(r) &= \mathrm{exp} \left( -\frac{1}{2} r^2 \right)\label{eq:matrix_kernels:rbf}\\
	k_\mathrm{Mat\acute{e}rn}^\mathcal{M}(r) &= \frac{2^{1-\nu}}{\Gamma \left( \nu \right)} \left( \sqrt{2\nu} r \right)^\nu K_\nu \left( \sqrt{2\nu} r \right)\label{eq:matrix_kernels:matern}
\end{align}
where $l$ is the length scale parameter, $\nu$ a parameter of the Matérn kernel and $K_\nu$ a modified Bessel function \cite{williams2006gaussian}.
As found by \citet{feragen2015geodesic}, this distance metric does not result in positive definite \gls{rbf} kernels for all choices of length scales.
As later found by the same authors \cite{feragen2016open}, for small enough length scales, positive definite kernels can be achieved which has been used by \citet{jaquier2020bayesian} to sample valid length scales.\footnote{At a computational and complexity cost, more involved kernels could ensure positive definiteness for all length scales \cite{jacquier2022geometry}.}
We determine valid length scales using this approach for our kernels and summarize them in \Cref{tab:matrix_kernels:length_scales}.

\end{nodiff}

\subsection{Diagonal Kernels}
\label{sec:matrix_kernels:diagonal}
To model \gls{dof}-aware kernels, we first propose to multiply a kernel with scalar output \eqref{eq:matrix_kernels:rbf} -- \eqref{eq:matrix_kernels:matern} with a diagonal kernel
\begin{equation}
	\bm{k}(\bm{s}, \bm{s}') = k(\bm{s}, \bm{s}') \cdot \bm{K}_{\mathrm{diag}}(\bm{s}, \bm{s}').
\end{equation}
$\bm{K}_\mathrm{diag}$ is a diagonal matrix with the $p$-th diagonal entry
\begin{equation}
	\bm{K}_{\mathrm{diag},p,p}(\bm{s}, \bm{s}') = k \left( \left| \left( \Log_{\bm{s}}^\mathcal{M} \left( \bm{s}' \right) \right)_p \right| \right).
	\label{eq:matrix_kernels:diagonal_kernel}
\end{equation}
Its diagonal elements are calculated from a kernel $k$ \eqref{eq:matrix_kernels:rbf}-\eqref{eq:matrix_kernels:matern} with input on $\mathbb{R}$ while all other entries are zero.
Positive definiteness of the resulting kernel is ensured through its construction using the Hadamard product $\odot$ of a transformation kernel~\cite{caponnetto2008universal} and the identity matrix $\bm{I}$ and the Kronecker product with a kernel $k(\bm{s}, \bm{s}')$ with scalar output \cite{carmeli2010vector}.

Note that this kernel is already, through the manifold logarithm $\Log_{\bm{s}}(\cdot)$ in \eqref{eq:matrix_kernels:diagonal_kernel}, expressed in $\mathcal{T}_{\bm{s}}\mathcal{M}$, thus not requiring parallel transport.
Among the manifolds considered in this work, we can formulate a valid kernel on $\mathbb{R}^n$, $\mathcal{S}^1$, $\mathcal{S}^3$ as well as their products, where $\Log_{\bm{s}}^{\mathcal{M}} \bm{s}' = -\Log_{\bm{s}'}^{\mathcal{M}} \bm{s}$ holds for our logarithm maps.
On $\mathcal{S}^2$, this property is not given due to the hairy ball theorem which would result in a non-positive definite and thus invalid kernel.
As before, valid length scales on different manifolds can be determined through sampling.

\subsection{Evaluation in Toy Examples}
\label{sec:matrix_kernels:experiments}
\begin{figure}
	\vspace{-1.5em}
	\centering
	\begin{subfigure}{0.49\columnwidth}
		\includegraphics[width=\columnwidth,trim={0 10 20 20},clip]{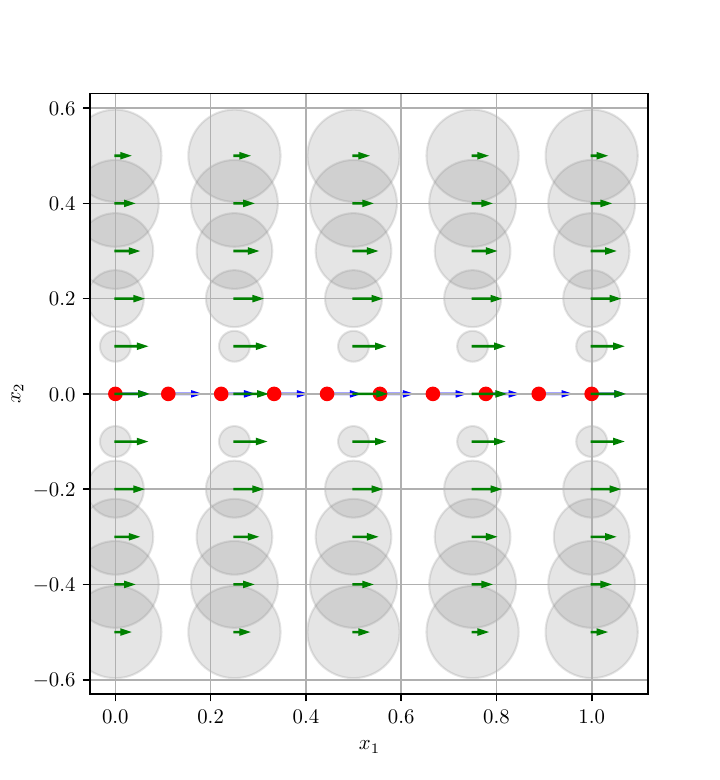}
	\end{subfigure}
	\begin{subfigure}{0.49\columnwidth}
		\includegraphics[width=\columnwidth,trim={0 10 20 20},clip]{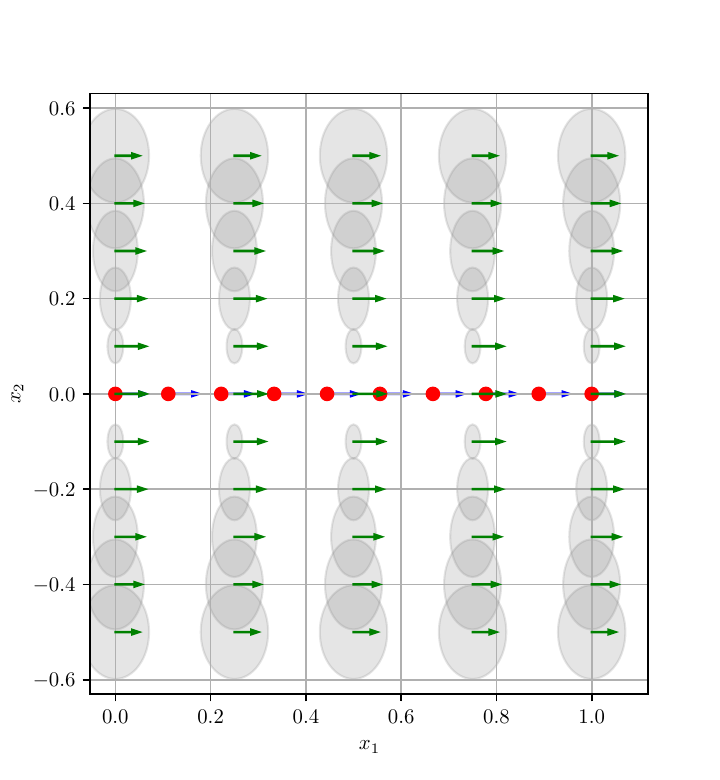}
	\end{subfigure}
	\begin{subfigure}{0.49\columnwidth}
		\includegraphics[width=\columnwidth,trim={0 10 20 42},clip]{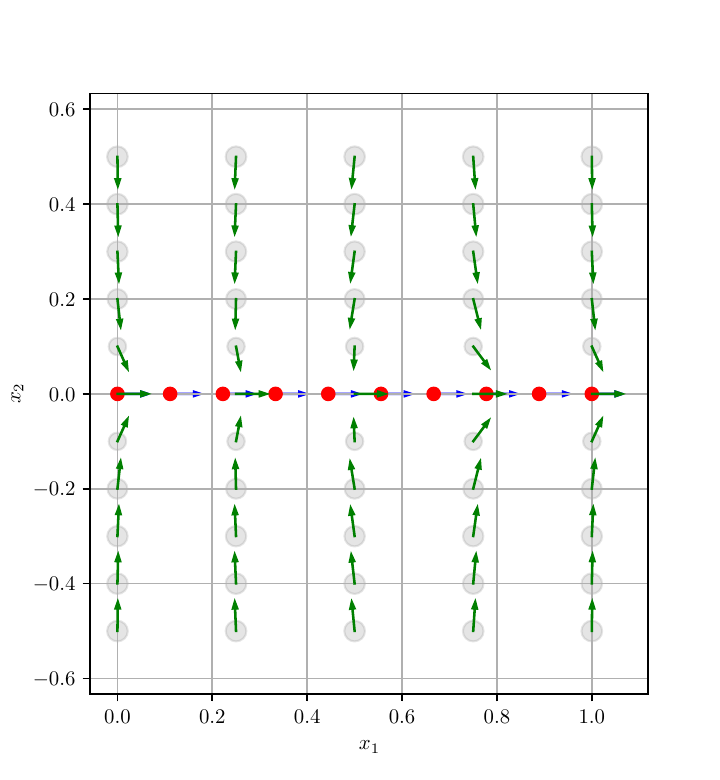}
	\end{subfigure}
	\begin{subfigure}{0.49\columnwidth}
		\includegraphics[width=\columnwidth,trim={0 10 20 42},clip]{kernel_diag_fused}
	\end{subfigure}
	\caption{\label{fig:matrix_kernels:toy} Upper row: \gls{kmp} output of a velocity policy with separable \gls{rbf} kernel ($l=0.3$ (left)) / diagonal \gls{rbf} kernel ($l_{RBF}=0.6$, $l_\mathrm{diag}=0.3$ (right)). Lower row: fusion of the \gls{kmp} with a \textit{stabilizing policy} \cite{muehlbauer2025unified}. Red dots visualize covariances and locations of known velocities (blue); grey ellipses the covariance and green arrows the velocity of \gls{kmp} and fusion, respectively.\vspace{-1em}}
\end{figure}
To show the usefulness of matrix-valued kernels, we learn velocity policies in 2D.
As depicted in \Cref{fig:matrix_kernels:toy}, points on $\bm{s} = \bm{x} \in \mathbb{R}^2$ (red) with known velocities on $\bm{\xi}(\bm{s}) = \dot{\bm{x}} \in \mathbb{R}^2$ (blue) are encoded in a \gls{kmp} using different kernels.
For the commonly used separable \gls{rbf} kernel (\Cref{sec:matrix_kernels:separable}), we only obtain scalar covariances; the kernel thus cannot output individual uncertainties per \gls{dof}.
Conversely, per-\gls{dof} uncertainties can be calculated by the diagonal kernel (\Cref{sec:matrix_kernels:diagonal}).
The covariance along the $x_2$ axis gets larger than along the $x_1$ axis when moving away from known points along that axis, highlighting the higher uncertainty for that \gls{dof}.

Following~\cite{muehlbauer2025unified}, those learned policies are fused with a \textit{stabilizing policy} with constant, diagonal covariance $\bm{\Sigma}_{\mathrm{stab}}$ and normalized velocity $\bm{\mu}_{\mathrm{stab}}$ pointing towards the closest point with known velocity.
As seen in  \Cref{fig:matrix_kernels:toy}, matrix-valued kernels allow the \textit{certain} component of a learned policy to be active for longer than with the separable kernel with identical entries on its diagonal.
This result is even more pronounced with more \glspl{dof} as used in \Cref{sec:evaluation:velocity} -- using such kernels allows the policy to proceed and only be corrected by the \textit{stabilizing policy} along \glspl{dof} with perturbations, given they are not too large, thus overcoming a central limitation in \cite{muehlbauer2025unified}.

\section{Computational Improvements}
\label{sec:computational_improvements}
Fast fitting to data and quick adaptations to novel situations are desirable properties of movement primitives.
In \glspl{kmp}, just as in \glspl{gp}, calculating the inverse required for mean \eqref{eq:kmp_mean_prediction} and covariance \eqref{eq:kmp_covariance_prediction} prediction is the computationally most expensive operation and additionally prone to numerical instabilities.
The Cholesky decomposition, as commonly used in \glspl{gp} \cite{williams2006gaussian,murphy2012machine}, avoids calculating explicit inverses and improves execution speed in \gls{kmp} predictions (\Cref{sec:computational_improvements:cholesky}) as well as in adding (\Cref{sec:computational_improvements:adding}) and removing (\Cref{sec:computational_improvements:removing}) measurements.
We adapt this approach to fitting, adding and removing points in our manifold-aware formulation and introduce a novel mechanism to restore the decomposition when removing points from the reference $\mathcal{D}$.

\subsection{Cholesky-factorized KMP (cKMP)}
\label{sec:computational_improvements:cholesky}
\begin{algorithm}[t]
\caption{\label{algo:ckmp:prediction} Prediction in a cKMP.}
\begin{algorithmic}
\State $\bm{L} \bm{L}^\top \gets \bm{K} + \lambda \bm{\Sigma}$ \Comment{Initial fitting \eqref{eq:cholfit}}
\State $\bm{\alpha} \gets \bm{L}^\top \backslash \bm{L} \backslash \Log_{\bm{\mu}}(\bm{e})$ \Comment{Pre-solve with \eqref{eq:cholsolve}}

\For{$\bm{s}_i^*$ in queries $\left[ \bm{s}_0^* \hdots \bm{s}_n^* \right]$}
	\State $\bm{\xi}(\bm{s}_i^*) \gets \mathrm{Exp}_{\bm{e}}\left(-\bm{k}^* \bm{\alpha} \right)$ \Comment{Predict mean \eqref{eq:kmp_mean_prediction}}
	\State $\bm{V} \gets \bm{L} \backslash \bm{k}^{*\top}$
	\State $\bm{\Sigma}(\bm{s}_i^*) \gets \bm{k}(\bm{s}_i^*, \bm{s}_i^*) - \bm{V}^\top \bm{V}$ \Comment{Predict covariance \eqref{eq:kmp_covariance_prediction}}
\EndFor
\end{algorithmic}
\end{algorithm}
Instead of calculating the inverse $\left( \bm{K} + \lambda \bm{\Sigma} \right)^{-1}$, we use the Cholesky factorization of the positive definite matrix $\bm{K} + \lambda \bm{\Sigma}$ into a product of lower triangular matrix $\bm{L}$ and its transpose
\begin{equation}
	\bm{L} \bm{L}^\top = \bm{K} + \lambda \bm{\Sigma}\label{eq:cholfit}
\end{equation}
which allows for solving the equation $\bm{L} \bm{L}^\top \bm{x} = \bm{b}$ through a forward and a backward substitution to obtain $\bm{x}$
\begin{gather}
	\bm{L} \bm{y} = \bm{b},\quad\bm{L}^\top \bm{x} = \bm{y}.\label{eq:cholsolve}
\end{gather}
\Cref{algo:ckmp:prediction} summarizes the mean and covariance prediction using Cholesky factorization, where we use $\backslash$ to denote the solving operation \eqref{eq:cholsolve} using forward- or backward substitution, resulting in a similar algorithm to that of predictions in a Cholesky-factorized \gls{gp} \cite{williams2006gaussian}, however adapted to observation data on manifolds through the use of $\Log \left( \cdot \right)$ and $\Exp \left( \cdot \right)$.
The computational complexity of the mean prediction is thus $\mathcal{O} \left( N M^2 \right)$, where $N$ is the size of $\bm{L}$ and $M$ the output dimension, and $\mathcal{O} \left( N^2 \right)$ for the covariance prediction dominated by the calculation of $\bm{V}$.
While $\bm{\alpha}$ can also be precomputed for the vanilla \gls{kmp} and iKMP prediction \eqref{eq:kmp_mean_prediction}, the default covariance prediction \eqref{eq:kmp_covariance_prediction} is $\mathcal{O} \left( 2 N^2 M \right)$.
This results in significantly improved prediction times of our Cholesky-factorized \gls{kmp} (cKMP) compared to \textit{vanilla} \gls{kmp} \cite{huang2019kernelized}.

\subsection{Appending Measurements}
\label{sec:computational_improvements:adding}
\begin{algorithm}[t]
\caption{\label{algo:ckmp:add_measurements} Adding measurements to a cKMP.}
\begin{algorithmic}
\For{$\bm{s}_i$, $\bm{\mu}_i$, $\bm{\Sigma}_i$ in \textit{new measurements}}
	\State $\bm{s} \gets [\bm{s}; \bm{s}_i]$, $\bm{\mu} \gets [\bm{\mu}; \bm{\mu}_i]$, $\bm{\Sigma} \gets [\bm{\Sigma}; \bm{\Sigma}_i]$ \Comment{append}
\EndFor
\State $\bm{K} \gets [\bm{K}; \bm{K}_{i, j}]$ \Comment{Calculate \& append new entries}
\State $\bm{L} \gets \mathrm{cholupdate}(\bm{K} + \lambda \bm{\Sigma})$ \Comment{through \eqref{eq:cholesky_decomp} -- \eqref{eq:cholesky_update_2}}
\end{algorithmic}
\end{algorithm}
\citet[B.1]{osborne2010bayesian} proposes an algorithm for adding rows and columns to a Cholesky factorization which we can readily use to calculate an updated $\bm{L}^*$ when appending new entries to the end of the matrix $\bm{A} = \bm{K} + \lambda \bm{\Sigma}$ we want to factorize.
Setting the original $\bm{A}$ as the upper left corner of $\bm{A}^*$
\begin{gather}
	\label{eq:cholesky_decomp}
	\bm{A}^* = \begin{pmatrix}
		\bm{A}^*_{1,1} & \bm{A}^*_{1,2}\\
		\bm{A}^*_{2,1} & \bm{A}^*_{2,2}
	\end{pmatrix},
	\quad
	\bm{L}^* = \begin{pmatrix}
		\bm{L}^*_{1,1} & \bm{0}\\
		\bm{L}^*_{2,1} & \bm{L}^*_{2,2}
	\end{pmatrix},
\end{gather}
where $\bm{A} = \bm{A}^*_{1,1}$, and through $\bm{A}^* = \bm{L}^* \bm{L}^{*\top}$, we obtain
\begin{align}
	\bm{L}^*_{1,1} = \bm{L},
	\quad
	\bm{L}^*_{1,1} \bm{L}^{*\top}_{2,1} = \bm{A}^*_{1,2},\label{eq:cholesky_update_1}\\
	\bm{L}^*_{2,2} \bm{L}^{*\top}_{2,2} = \bm{A}^*_{2,2} - \bm{L}^*_{2,1} \bm{L}^{*\top}_{2,1}.\label{eq:cholesky_update_2}
\end{align}
We can thus reuse the precomputed $\bm{L}$ as upper left block $\bm{L}_{1,1}^*$, $\bm{L}^*_{2,1}$ can then be obtained using forward substitution and $\bm{L}^*_{2,2}$ through a Cholesky decomposition.

The full algorithm is denoted in \Cref{algo:ckmp:add_measurements}.
As we add measurements to the end of the decomposition, the complexity of \eqref{eq:cholesky_update_1} is $\mathcal{O}(N^2)$ where $N$ is the size of the original $\bm{L}$ and the complexity of \eqref{eq:cholesky_update_2} is $\mathcal{O}(D^3)$ where $D$ is the dimension of the measurements to be added.
This results in a clear computational advantage over the naive recomputation of the inverse used by the original \gls{kmp} and is also significantly faster than iKMP (\Cref{tab:computational_improvements:letter,tab:computational_improvements:trajectory}).

\subsection{Removing Measurements}
\label{sec:computational_improvements:removing}
For removing a measurement, we permute $\bm{K} + \lambda \bm{\Sigma}$ with a permutation matrix $\bm{P}$ to obtain
\begin{equation}
	\bm{P} \left( \bm{K} + \lambda \bm{\Sigma} \right) \bm{P}^\top = \bm{P} \bm{L} \bm{L}^\top \bm{P}^\top = \bm{L}^{'} \bm{L}^{'\top}.
\end{equation}
Note that in case we only want to remove measurements at the end (which is often required when we iteratively update a via point) $\bm{P} = \bm{I}$ and we can simply remove the last rows and columns of $\bm{L}$, $\bm{K}$ and $\bm{\Sigma}$ to obtain $\bm{L}^{'}$, $\bm{K}^{'}$ and $\bm{\Sigma}^{'}$, resulting in a negligible computational cost.
In case we want to remove a measurement before the end, the lower-triangular form of $\bm{L}^{'}$ has to be restored as shown in \Cref{sec:appendix:rediag}.

\subsection{Evaluation in Toy Examples}
\label{sec:computational_improvements:adding:experiments}
\begin{figure}
	\centering
	\begin{subfigure}{0.49\columnwidth}
		\includegraphics[width=\columnwidth,trim={20 10 47 35},clip]{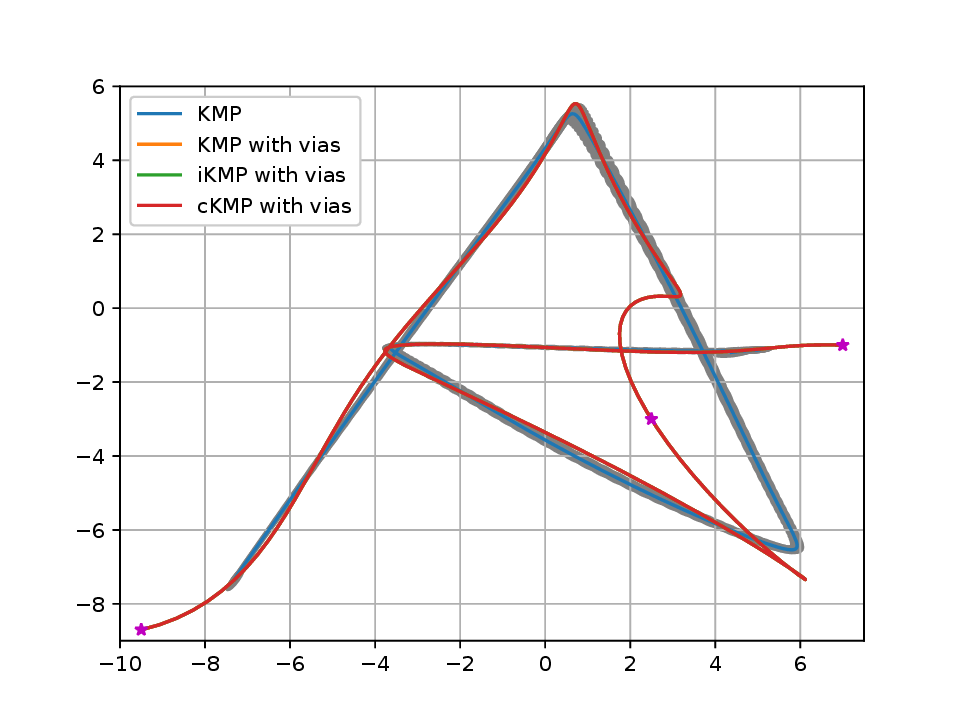}
	\end{subfigure}
	\begin{subfigure}{0.49\columnwidth}
		\includegraphics[width=\columnwidth,trim={105 45 65 45},clip]{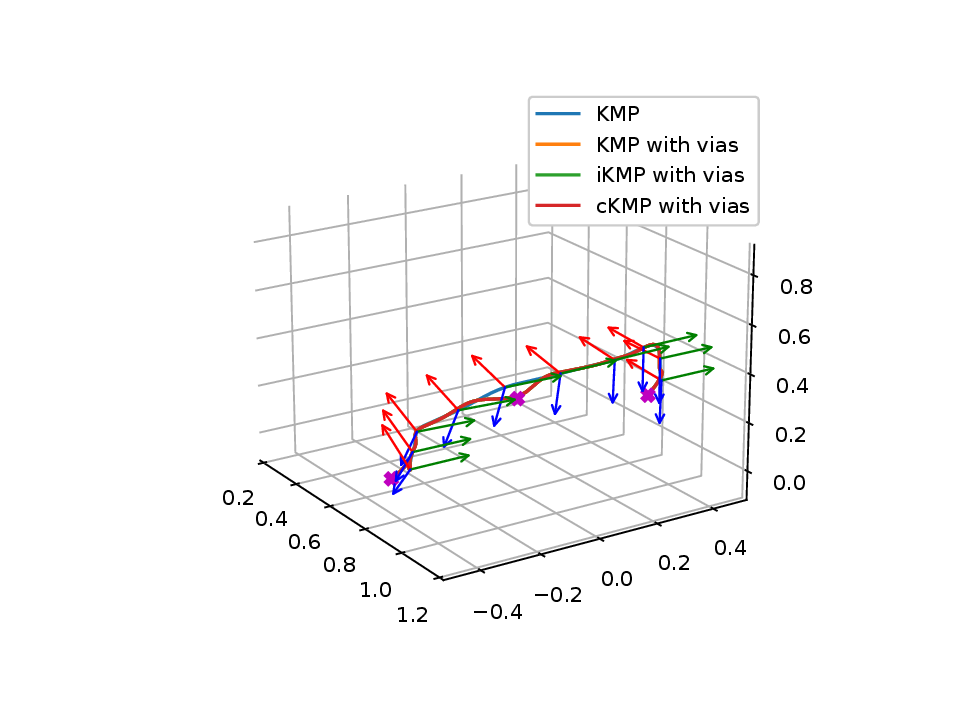}
	\end{subfigure}
	\caption{\label{fig:computational_improvements:letter} Adapting the letter ``A'' in the letter writing task \cite{silverio2023nonparametric} (left) and a full trajectory of $500$ poses on $\mathcal{M}_1$ (right; via points shown as magenta $\bm{x}$). Plots for different calculation methods are indistinguishable as they result in the same trajectory, proving their numerical equivalence.}
\end{figure}
\begin{table}
\scriptsize
\caption{\label{tab:computational_improvements:letter}\centering \scshape Computation times (milliseconds) for adding and removing $3$ points and predicting $200$ points for letter writing.}
\centering
\begin{tabular}{@{}c|cccc@{}}
Method & Fit (\ref{sec:computational_improvements:cholesky}) & Add (\ref{sec:computational_improvements:adding}) & Remove (\ref{sec:computational_improvements:removing}) & Predict (\ref{sec:computational_improvements:cholesky})\\
\hline
N.-W.~\cite{lopez2023non} & $0.12 \pm 0.00$ & -/- & -/- & $0.83 \pm 0.72$\\
\gls{kmp} \cite{huang2019kernelized} & $12.93 \pm 3.93$ & $13.63 \pm 3.97$ & $11.42 \pm 3.16$ & $1.68 \pm 0.33$\\
iKMP \cite{wang2026computationally} & $12.17 \pm 3.01$ & $1.18 \pm 0.09$ & $0.84 \pm 2.33$ & $1.71 \pm 0.58$\\
cKMP (\ref{sec:computational_improvements}) & $\bm{2.66 \pm 0.78}$ & $\bm{0.82 \pm 0.16}$ & $\bm{0.60 \pm 0.58}$ & $\bm{1.35 \pm 0.59}$
\end{tabular}
\vspace{-1em}
\end{table}

\begin{table}
\scriptsize
\caption{\label{tab:computational_improvements:trajectory}\centering \scshape
Computation times (milliseconds) fitting, adding $3$ poses and subsequently removing them in a trajectory with $100$ reference poses, predicting $500$ poses.}
\centering
\begin{tabular}{@{}c|cccc@{}}
Method & Fit (\ref{sec:computational_improvements:cholesky}) & Add (\ref{sec:computational_improvements:adding}) & Remove (\ref{sec:computational_improvements:removing}) & Predict (\ref{sec:computational_improvements:cholesky})\\
\hline
N.-W.~\cite{lopez2023non} & $0.22 \pm 0.01$ & -/- & -/- & $24.06 \pm 2.89$\\
R. \gls{kmp} (\ref{sec:kmp}) & $37.11 \pm 3.74$ & $37.20 \pm 1.72$ & $34.51 \pm 2.61$ & $13.28 \pm 1.22$\\
iKMP \cite{wang2026computationally} & $36.22 \pm 3.35$ & $4.11 \pm 1.60$ & $2.29 \pm 0.52$ & $13.15 \pm 0.95$\\
cKMP (\ref{sec:computational_improvements}) & $\bm{7.89 \pm 2.32}$ & $\bm{2.64 \pm 0.32}$ & $\bm{1.27 \pm 0.37}$ & $\bm{8.57 \pm 0.82}$
\end{tabular}
\end{table}

We provide computation times for the original \gls{kmp} formulation, iKMP \cite{wang2026computationally} and the Cholesky-factorized \gls{kmp} (cKMP) introduced in this section for initial fitting (\textit{vanilla} \gls{kmp} / iKMP: calculation of the inverses, cKMP: Cholesky decomposition, both: calculation of $\bm{\alpha}$ (\Cref{algo:ckmp:prediction})),  adding of $3$ via points and subsequent removal of those as well as the full prediction for letter writing in \Cref{tab:computational_improvements:letter} and for a full trajectory prediction on $\mathcal{M}_1$ (\Cref{sec:eval:adapting}) in \Cref{tab:computational_improvements:trajectory} (both shown in \Cref{fig:computational_improvements:letter}).%
\footnote{Timings and their standard deviations were computed for $100$ executions on a computer with Intel Core i7-10700K CPU using Ubuntu 24.04.
The computations were performed and measured in parallelized C++ code compiled with AVX optimizations enabled.}
As baseline, we compare with a Riemannian version of Nadaraya-Watson regression (\cite{muehlbauer2026geometric}) inspired by~\cite{lopez2023non} -- note that while this baseline compares favorably on small problems, for movements with many points, it quickly gets much slower as it takes every single data point into account at prediction time.
This method furthermore can only provide an \textit{aleatoric} uncertainty estimate, but no \textit{epistemic} uncertainty.
Furthermore, via points can only be added by using custom engineered kernels as suggested by \cite{lopez2023non}; we have omitted their implementation.
Adding nullspace modifications to the \gls{kmp} prediction does not measurably increase computation times; timings are thus omitted from \Cref{tab:computational_improvements:letter,tab:computational_improvements:trajectory}.

Initial fitting, adding and removing points in the original \gls{kmp} formulation take approximately the same time, as a full inversion of $\bm{K} + \lambda \bm{\Sigma}$ is required.
Even though we also precompute $\bm{\alpha}$ (\Cref{algo:ckmp:prediction}) for the \textit{vanilla} \gls{kmp} and iKMP, our proposed cKMP is still significantly faster due to a quicker covariance calculation.
Note that for adding and removing points in the original \gls{kmp}, already reallocating smaller or bigger memory areas for $\bm{K}$ and $\bm{\Sigma}$ takes considerable time as each matrix with $(D \cdot N)^2 = 600^2$ entries for $D=6$, $N=100$ in double precision occupies \SI{2.88}{\mega\byte} of memory.
In iKMP, only the inverse and in cKMP, only the Cholesky decomposition have to be updated while full $\bm{K}$ and $\bm{\Sigma}$ are only calculated for initial fitting.

\section{Evaluation on a Robotic System}
\label{sec:evaluation}
We evaluate our approach in real experiments that either would not have been possible before or benefit significantly from our contributions.
We use the on-manifold impedance and velocity controllers of~\cite{muehlbauer2025unified}.
All experiments use the Riemannian \gls{kmp} prediction (\Cref{sec:kmp:prediction}).
The first two experiments (\Cref{sec:eval:tp,sec:eval:tp_comparison}) furthermore employ task parameterization (\Cref{sec:kmp:task_parameterization}), the third experiment (\Cref{sec:evaluation:velocity}) demonstrates the usefulness of matrix-valued kernels (\Cref{sec:matrix_kernels}) in \glspl{ds}.
In \Cref{sec:eval:adapting}, we use the movement primitive adaptation (\Cref{sec:kmp:adaptation}).
This experiment also profits exceptionally from the computational improvements (\Cref{sec:computational_improvements}), allowing for a continuous trajectory update.
The final experiment in \Cref{sec:eval:missing_data} highlights the importance of an epistemic uncertainty estimate.
We show the execution of all experiments in the supplementary video.

\subsection{Task Parameterization on Different Manifolds}
\label{sec:eval:tp}
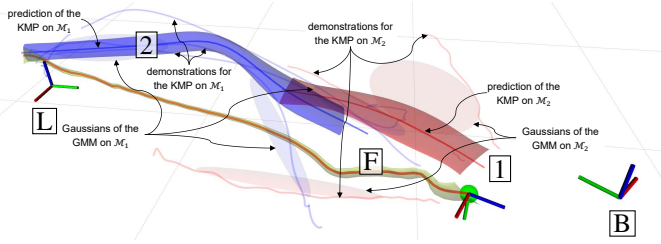
\begin{figure}
	\centering
	\annotategraphicsmulti{
		\includegraphics[width=\columnwidth,trim={0 60 0 90},clip,page=5]{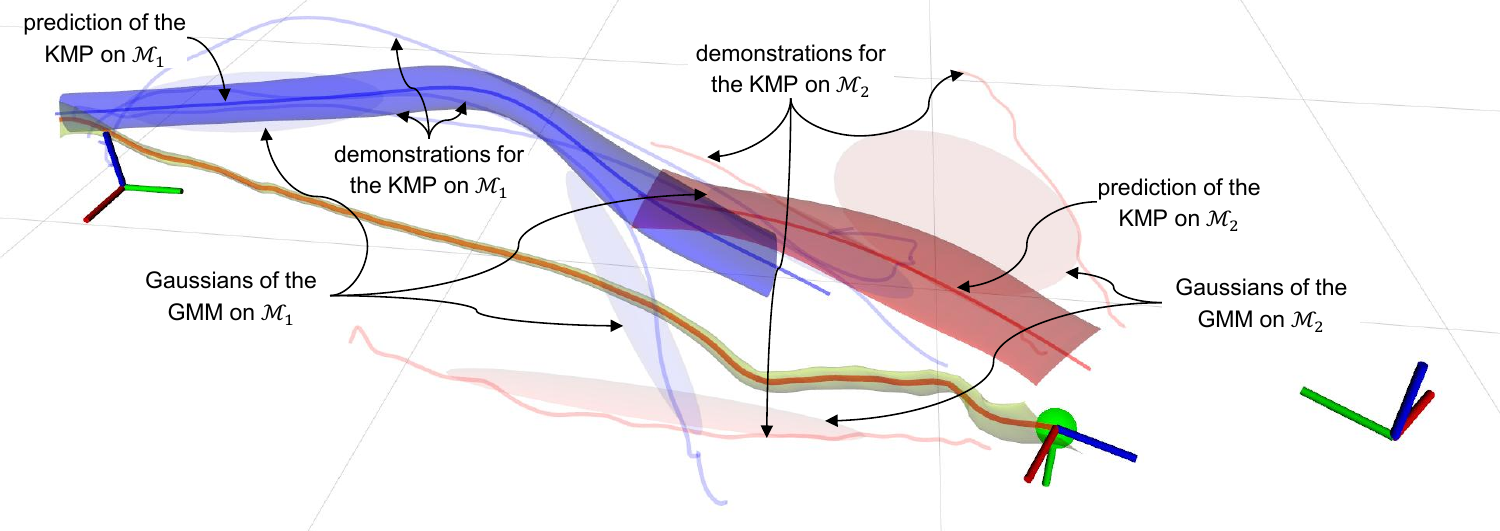}
	}{
	    \node at (0.92,0.07) {B};
	    \node at (0.07,0.5) {L};
	    \node at (0.74,0.3) {1};
	    \node at (0.22,0.82) {2};
	    \node at (0.55,0.33) {F};
    }
	\caption{\label{fig:eval:bottle} Riemannian task-parameterized \gls{kmp} on $\mathcal{M}_2$ for picking up a bottle from any angle at \protect\lref{B} and placing it at \protect\lref{L}. The model~\protect\lref{1} on $\mathcal{M}_2$ (red demonstrations / Gaussians / covariance; origin \protect\lref{B}) is fused with a model~\protect\lref{2} on $\mathcal{M}_1$ (blue demonstrations / Gaussians / covariance; origin \protect\lref{L}) to a model~\protect\lref{F} on $\mathcal{M}_2$ (yellow covariance).\vspace{-1em}}
\end{figure}

We first evaluate task parameterization on $\mathcal{M}_1$ and $\mathcal{M}_2$ (\Cref{sec:kmp:task_parameterization}) on the task of grasping a bottle~\protect\lref{B}, which is a rotationally symmetric problem, and placing it at a specified location~\protect\lref{L} (\Cref{fig:eval:bottle}).
The \gls{kmp} \protect\lref{1} is learned from three demonstrations of picking up the bottle from different angles at \protect\lref{B}.
By using $\mathcal{M}_2$ centered at \protect\lref{B}, this results in a high covariance along the angular \gls{dof} which we use to modulate a low stiffness.
The \gls{kmp} \protect\lref{2} on $\mathcal{M}_1$ is learned from $3$ demonstrations of placing the bottle at \protect\lref{L} in a relatively constrained manner, thus resulting in lower covariance and stronger guidance along the angular \gls{dof} when fused on $\mathcal{M}_2$ depicted as \protect\lref{F}.
From this combination of policies, the operator can pick the bottle \protect\lref{B} from different angles, while the placement at \protect\lref{L} is constrained.
The \textit{aleatoric} uncertainty plays a key role in this experiment -- both \glspl{kmp} are queried in a region with available data and thus low \textit{epistemic} uncertainty.
The uncertainty associated with those known data points accordingly determines both the fused result and the stiffness.
The supplementary video shows the resulting impedance behavior (stabilized using \cite{muehlbauer2026stabilizingarxiv}) of the robot along the complete task.

\subsection{Comparison to Vanilla Task-Parameterized KMPs}
\label{sec:eval:tp_comparison}
\begin{figure}
	\centering
	\begin{subfigure}{0.49\columnwidth}
		\includegraphics[width=\columnwidth,trim={20 150 20 310},clip]{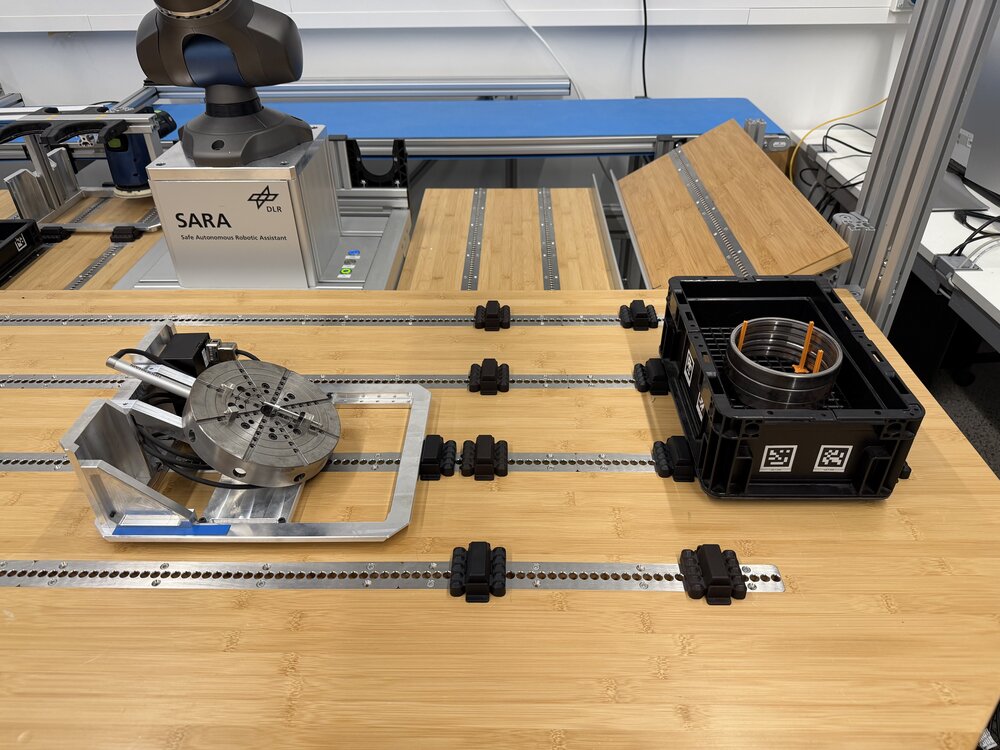}
	\end{subfigure}
	\begin{subfigure}{0.49\columnwidth}
		\includegraphics[width=\columnwidth,trim={20 200 20 260},clip]{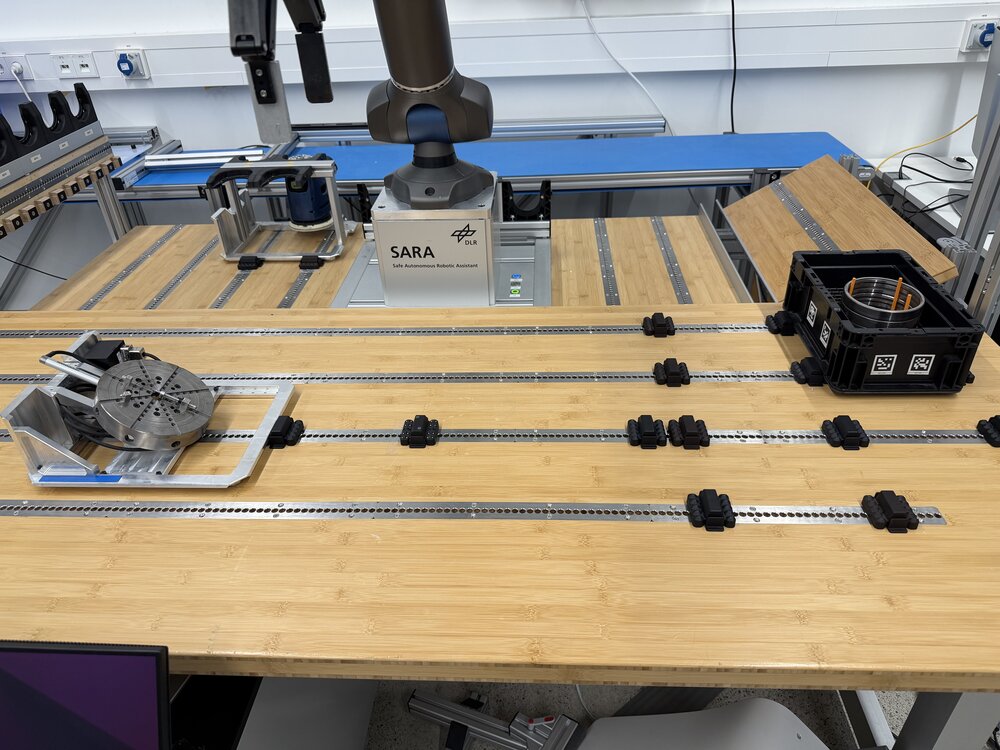}
	\end{subfigure}
	\vspace{-0.9em}

	\begin{subfigure}{0.49\columnwidth}
		\includegraphics[width=\columnwidth,trim={20 240 20 220},clip]{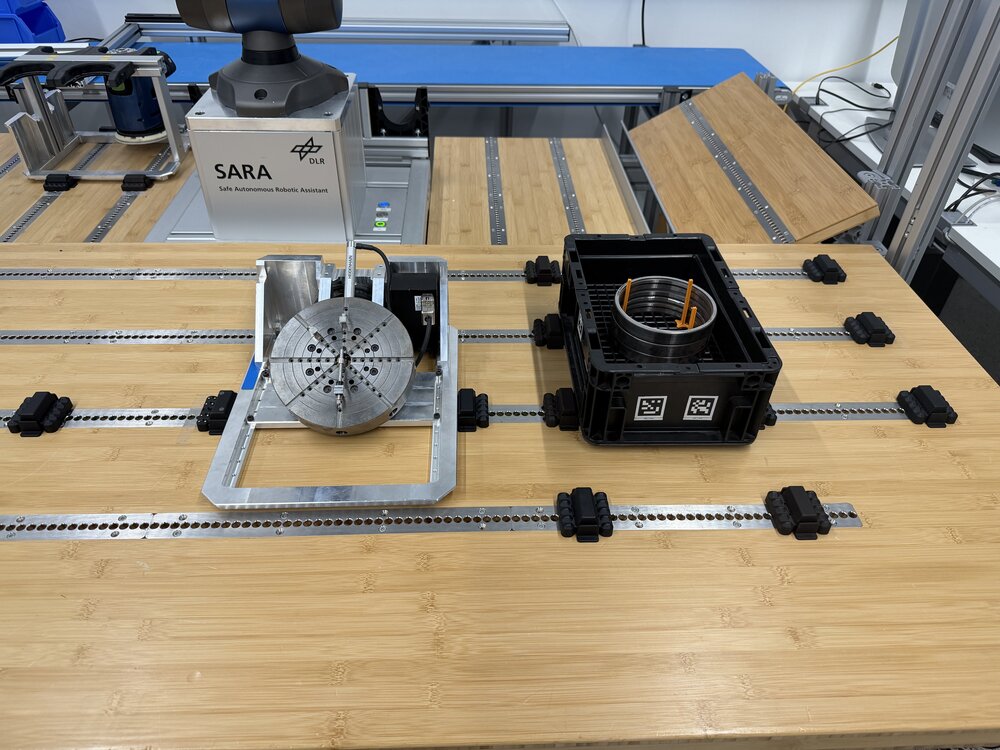}
	\end{subfigure}
	\begin{subfigure}{0.49\columnwidth}
		\includegraphics[width=\columnwidth,trim={20 160 20 300},clip]{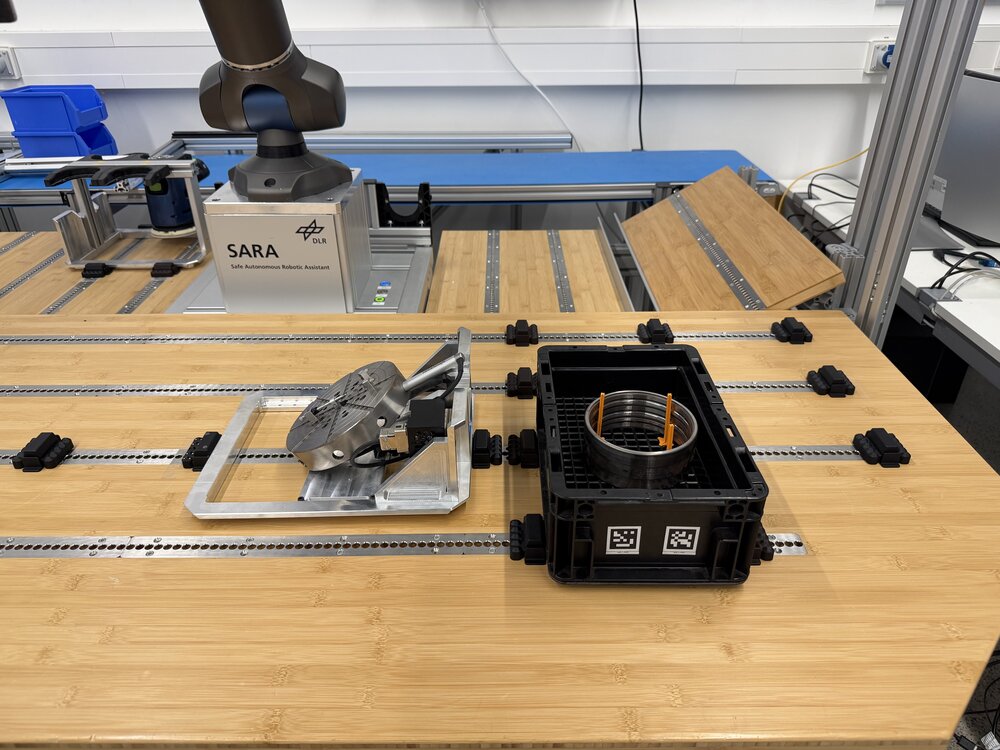}
	\end{subfigure}
	\caption{\label{fig:tp_comparison_variants} Experimental setup for grasping bearing rings from a box and inserting them in a measurement station. The station is rotated at \SI{0}{\degree} (upper row) with two different offsets and at \SI{90}{\degree} and \SI{180}{\degree} (lower row). While rotations of \SI{0}{\degree} and \SI{90}{\degree} are used for training the task-parameterized model on $\mathcal{M}_1$, the rotation of \SI{180}{\degree} is only seen at test time.\vspace{-1em}}
\end{figure}
We compare our task-parameterized \gls{kmp} on the measurement ring insertion task from \cite{knauer2026irosa,knauer2026clasp} using their Open-Source implementation\footnote{\url{https://github.com/DLR-RM/interactive-incremental-learning}} with both the vanilla task-parameterized \gls{kmp} available in the code and our Riemannian implementation on $\mathcal{M}_1$.
\Cref{fig:tp_comparison_variants} shows the setup used for training and evaluation.
We use $15$ demonstrations in total, $10$ for the \SI{0}{\degree} and $5$ for \SI{90}{\degree} measurement station orientations shown therein, corresponding to the possible locations of the black box.
\textit{Task frames} are recorded as the end effector location when closing the gripper for picking up the ring respectively when opening the gripper for releasing it.
We train two \glspl{kmp}, one for picking and one for placing the ring in the respective coordinate frames.
At test time, these task frames are set to box and measurement station locations to allow for generalization.

As seen in the supplementary video, while the vanilla implementation succeeds when it is only trained with one orientation, as soon as samples with the measurement station rotated at \SI{90}{\degree} are added, the orientation prediction fails.
This happens as treating quaternions as vector-valued data in the original \gls{kmp} neglects their geometry, such that even after renormalization, the obtained orientation is meaningless.
Contrarily, our approach succeeds in the insertion task for all possible combinations, providing smooth trajectories also for the case of the measurement station rotated by \SI{180}{\degree} which has not been part of the training data.

\subsection{Dynamical Systems with Matrix-valued Kernels}
\label{sec:evaluation:velocity}
\begin{figure}
	\centering
	\annotategraphicsmulti{
		\includegraphics[width=\columnwidth,trim={0 0 0 70},clip,page=3]{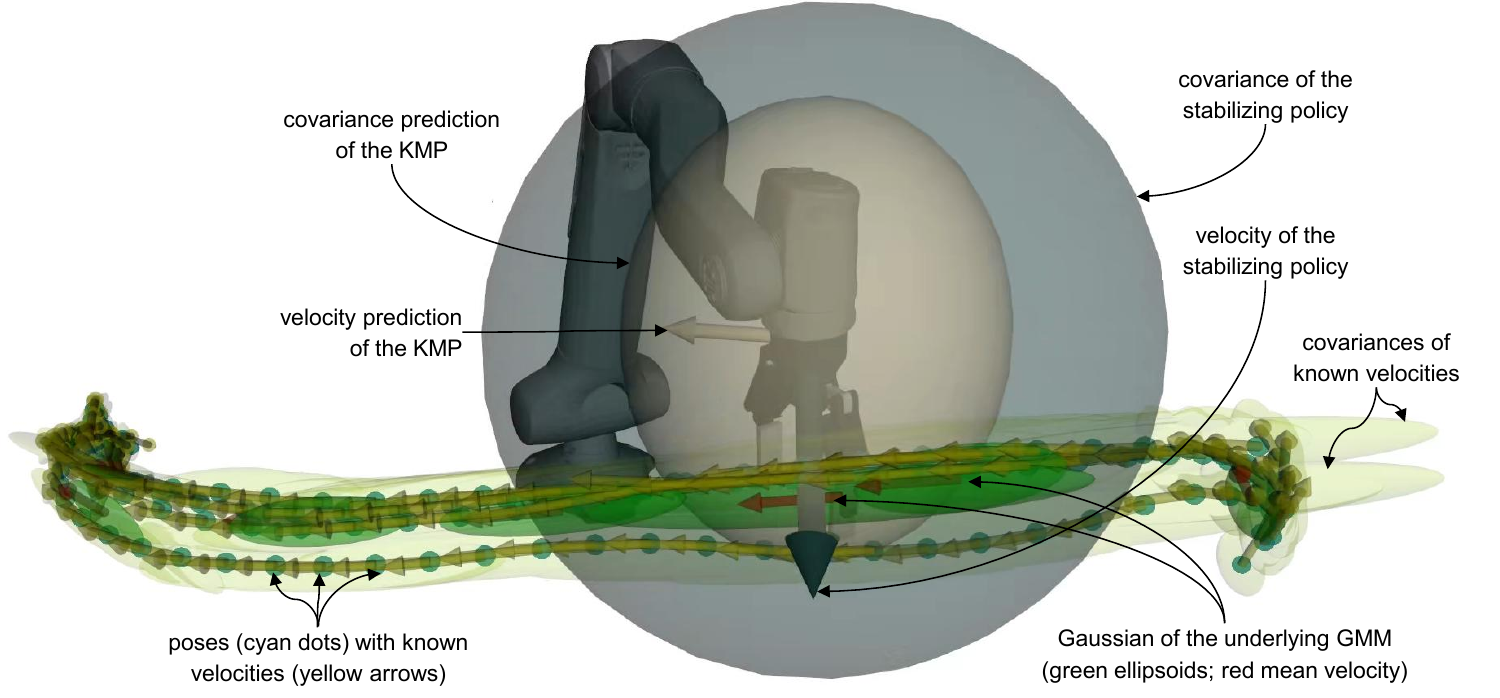}
	}{
	    \node at (0.9,0.17) {C1};
	    \node at (0.07,0.5) {C2};
    }
	\caption{\label{fig:eval:vel_policy} Riemannian \gls{kmp} with diagonal Matérn kernel for a state-based policy ($\bm{s} = \bm{x} \in \mathcal{M}_1$, $\bm{\xi}(\bm{s}) = \dot{\bm{x}} \in \mathcal{T}_{\bm{x}}\mathcal{M}_1$) to move a ring from \protect\lref{C1} to \protect\lref{C2}. Note the higher covariance of the \gls{kmp} prediction around the perturbed $z$ axis, leading to a correction by the \textit{stabilizing policy} only along this \gls{dof}.\vspace{-1em}}
\end{figure}
\begin{figure}
	\centering
	\includegraphics[width=\columnwidth,trim={0 0 0 0},clip]{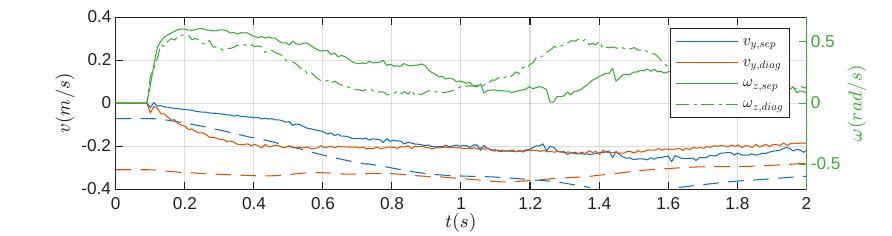}
	\caption{\label{fig:eval:vel_perturbation} Velocities in $y$ (measured: solid; desired: dashed) and around the $z$ axis using a separable (subscript \textit{sep}) and a diagonal Matérn kernel (subscript \textit{diag}) when starting with a rotational perturbation around the $z$ axis at $t=0$.\vspace{-1em}}
\end{figure}
To evaluate matrix-valued kernels (\Cref{sec:matrix_kernels}), we encode a velocity policy for moving a ring from \protect\lref{C1} to \protect\lref{C2} (\Cref{fig:eval:vel_policy}) learned from four kinesthetic demonstrations in a \gls{kmp} using both a separable Matérn kernel with $\nu = \frac{3}{2}$ and $l = \left( 0.15, 0.15, 0.15, 0.2, 0.2, 0.2 \right)$ and a diagonal Matérn kernel using the same parameters for the diagonal part and $l = \left( 0.4, 0.4, 0.4, 0.8, 0.8, 0.8 \right)$ for the scalar kernel.
The learned policy (\Cref{fig:eval:vel_policy}) is fused with a \textit{stabilizing policy}~\cite{muehlbauer2025unified} pointing towards the closest known pose with velocity $\mu_\mathrm{stab} = \SI{0.5}{\metre\per\second}$ and $\bm{\Sigma}_{\mathrm{stab}} = \SI{0.1}{\metre\squared} \cdot \bm{I}$ \cite[(26)]{muehlbauer2025unified}.

While both kernels result in similar motions for the unperturbed case, their perturbed executions differ.
We start the policy at $t=0$ in the middle of the trajectory with a perturbation of \SI{-0.7}{\radian} around the $z$ axis.
As seen in \Cref{fig:eval:vel_perturbation} and the supplementary video, while the \textit{stabilizing policy} corrects the rotation around the $z$ axis similarly, the system accelerates more quickly with the diagonal kernel along the $y$ axis -- the main motion direction.
This is possible as the diagonal kernel can model per-\gls{dof} \textit{epistemic} covariances during perturbations, thus allowing the policy to control the motion along $y$ earlier than in the separable case.

Note that beyond these differences, the experiment in this section is enabled by the \textit{epistemic uncertainty} estimate of our approach -- e.g., \glspl{gmm} which do only output \textit{aleatoric uncertainty} cannot model distance to known data.
The \textit{stabilizing policy} would thus never get activated, leading to ever increasing errors in case of perturbations.

\subsection{Adapting Kernelized Trajectory Fixtures}
\label{sec:eval:adapting}
\begin{figure}
	\vspace{-1em}
	\centering
	\includegraphics[width=\columnwidth,trim={0 0 230 185},clip,page=1]{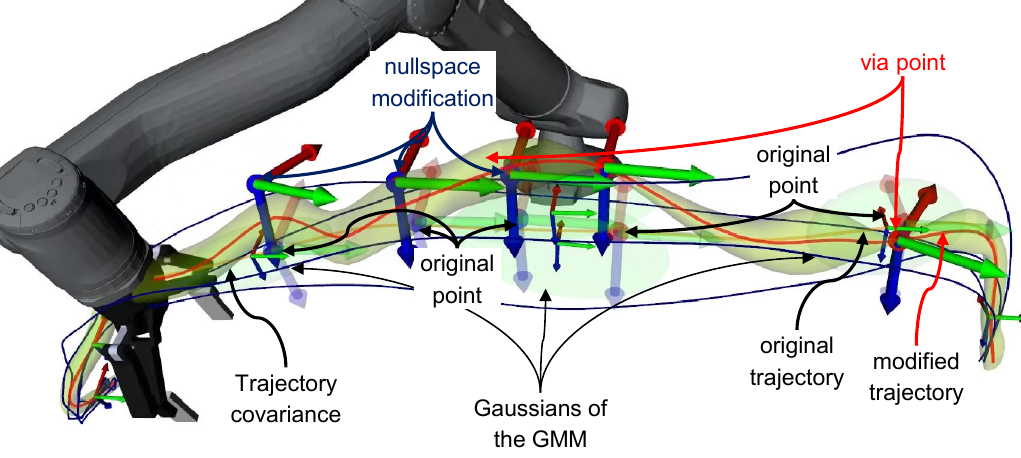}
	\caption{\label{fig:eval:adaptation} Real-time adaptation of a trajectory with different modifications.\vspace{-1em}}
\end{figure}
Next, we show the adaptation of a time-based trajectory ($\bm{s} \in \mathbb{R}$, $\bm{\xi}(\bm{s}) \in \mathcal{M}_1$) learned from $4$ demonstrations using via points and null space adaptations (\Cref{sec:kmp:adaptation}) in the same experimental run of a shared control scenario.
Both modification types are compatible and can be chosen according to task requirements.
While we physically interact with the robot, a teleoperated coupling would also be possible.
The task to be performed by a human operator guided by a trajectory (\Cref{fig:eval:adaptation}) is to pick a ring from a container and place it on the measurement station while avoiding an unmodeled obstacle.
We use a Matérn kernel ($\nu=\frac{5}{2}$, $l = 0.3$) for the \gls{kmp}, a covariance of $\bm{\Sigma} = \SI{1e-8}{\metre\squared}$ for via points and decorrelate null space actions using the approach introduced in \cite{quere2024probabilistic} to achieve more intuitive modifications.
This experiment is performed with the end effector rotated by \SI{180}{\degree} around the $y$-axis -- the setting where the \gls{sts} \gls{kmp} fails in \Cref{sec:kmp:experiments}.

The user can start a modification by clicking a button on the robot.
Depending on the selected mode, either a new via point or a new null space modification with the current end effector pose is applied at the current $\bm{s}$ of the trajectory.
This modification is continuously updated by removing the previous modification and applying a new one at the new end effector pose while preserving previous via points and null space actions.
Using the computational improvements introduced in \Cref{sec:computational_improvements}, calculating the update and updating the trajectory in the controller for $100$ extracted poses takes less than \SI{10}{\milli\second} compared to \SI{40}{\milli\second} without.
We use a constant update rate of \SI{10}{\hertz}, while fading between subsequent trajectories in the controller over $\Delta t =\SI{0.5}{\second}$ from $\bm{x}_0$ to $\bm{x}_1$ along the Riemannian geodesic.
This update rate is chosen deliberately at just \SI{2}{\hertz} in order not to close the control loop through the adaptation, avoiding potential instabilities.
A modification to circumvent an obstacle is shown in \Cref{fig:eval:adaptation}, where $2$ via points and $3$ null space modifications have been applied to correct both position and orientation.
While via points achieve a \textit{precise} modification, with the trajectory passing directly through them, they also lower the predicted uncertainty.
Contrarily, null space modifications are more \textit{coarse} and depend on and preserve the learned covariance -- larger modifications are thus required in the lower covariance region on the left to achieve the same modification as in the middle.

\subsection{Kernelized Trajectory Fixtures With Missing Data}
\label{sec:eval:missing_data}
\begin{figure}
	\vspace{-1em}
	\centering
	\includegraphics[width=\columnwidth,trim={0 35 0 38},clip,page=6]{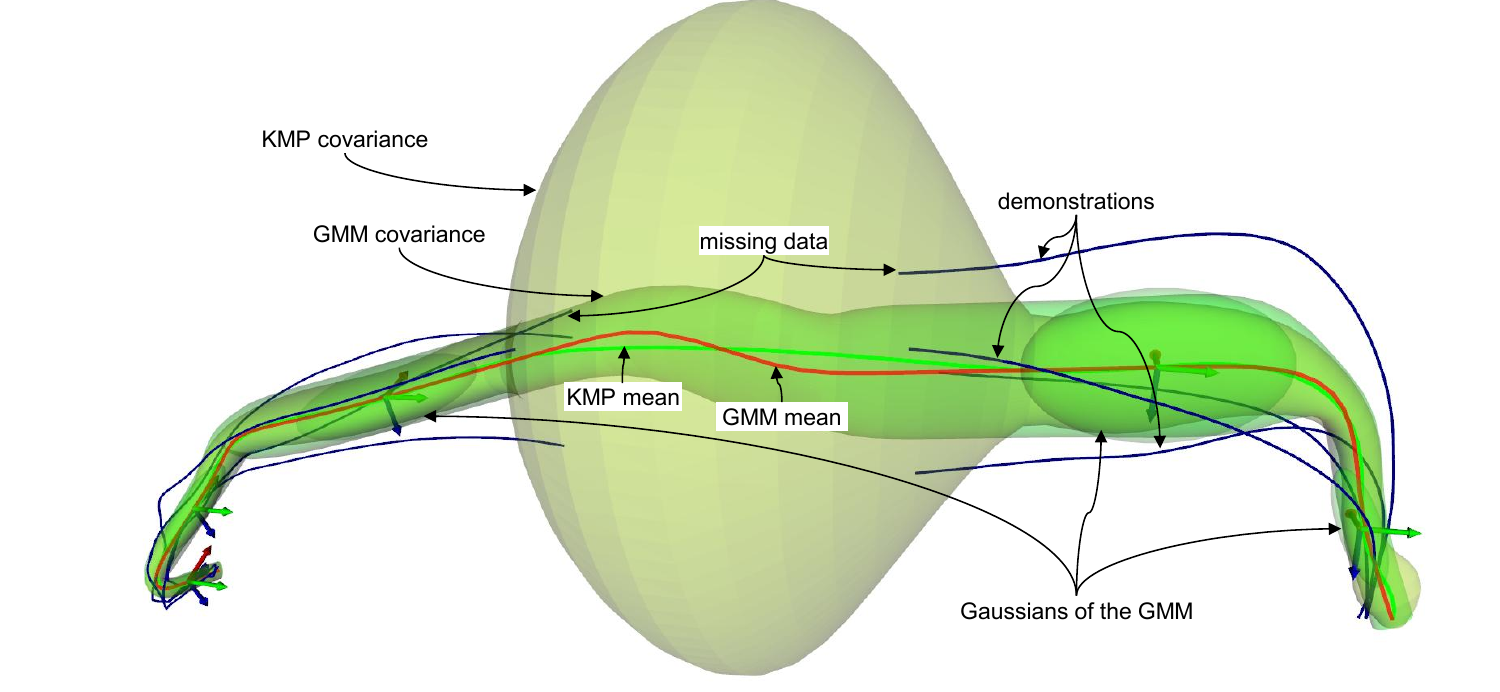}
	\caption{\label{fig:eval:missing_data} Comparison of Riemannian \gls{gmm} and \gls{kmp} predictions with missing data. Missing demonstration data, signified by cut outs of the demonstration trajectories, leads to increased covariance of the \gls{kmp} prediction while the \gls{gmm} incorrectly predicts high certainty.\vspace{-1em}}
\end{figure}
To evaluate the significance of the epistemic uncertainty prediction of our approach, we compare with a \gls{gmm} as baseline.
As seen in \Cref{fig:eval:missing_data}, the \gls{gmm} incorrectly predicts low uncertainty in regions of missing data while the epistemic uncertainty captured by our approach predicts an increased uncertainty.
This covariance prediction can be used by downstream algorithms \cite{muehlbauer2025unified,muehlbauer2026stabilizingarxiv} to reduce the stiffness of the robot in uncertain regions as shown in the supplementary video.

\section{Discussion}
\label{sec:limitations}
Our work leverages the expressiveness of Gaussian distributions on different Riemannian manifolds $\mathcal{M}$.
By modeling the motion on a manifold including the orientation, as opposed to a purely Euclidean modeling, we can represent orientations and support the full range of possible rotations.
Depending on the manifold, e.g., rotationally symmetric tasks can be learned and fused efficiently using our extended task parameterization.

We believe our method is best used in scenarios where explainability and fast computations are required, such as in shared control with humans in the loop.
Especially applications with few available demonstrations, a need for fast adaptations and high precision requirements benefit from our approach.
Ensuring manifold-valued and reasonable outputs, our method output can directly be fed to controllers without additional safety layers, as demonstrated in the robotic experiments.

We have successfully demonstrated operations with up to six dimensions; higher dimensional input like images would, however, require preprocessing such as task frame extraction.
Finally, generalization of the model is limited to areas close to available data -- further away, the prediction converges to the process mean $\bm{e}$, while the model outputs a high epistemic uncertainty that can, e.g., be used to activate a \textit{stabilizing policy}.

\section{Conclusion}
\label{sec:conclusion}
We have presented a Riemannian approach to probabilistic non-parametric movement modeling, providing both aleatoric and epistemic uncertainties.
The on-manifold formulation allows us to model motions with arbitrary orientation changes, without being limited to an area around a linearization point.
Through an extended task parameterization scheme, the properties of different manifolds can be leveraged.
We overcome the limitation of commonly used separable kernels which are only able to predict a scalar-valued epistemic uncertainty by proposing a novel diagonal kernel, which allows us to better model state-based policies.
Finally, through computational improvements by using the Cholesky decomposition and by using two adaptation schemes, trajectory modification, e.g., through kinesthetic human input, is possible in real time.
While both non-separable kernels and computational improvements could also be applied to the Euclidean \gls{kmp}, they are even more important when learning additional \glspl{dof}.
In future work, we plan to leverage additional matrix-valued kernels, e.g., curl-free kernels \cite{alvarez2012kernels} which can also predict off-diagonal epistemic uncertainty and to showcase our approach on discrete geometries.



\bibliographystyle{IEEEtranN}
\bibliography{literatur}

\begin{appendices}
\section{Function-space Derivation of Riemannian KMPs}
Alternatively to the weight-space derivation of Riemannian \glspl{kmp} in \Cref{sec:kmp:prediction}, we can also derive them, similarly to \glspl{gp}, in function space to show the similarities to Riemannian Gaussian Conditioning.
To this end, we first revisit Gaussian Conditioning~\cite{zeestraten2017manifold} in \Cref{sec:appendix:gaussian_conditioning} using an on-manifold optimization scheme.
We then show that Riemannian \glspl{kmp} can be formulated similarly, leading to the same result as before in \Cref{sec:appendix:prediction_conditioning}.

\subsection{On-Manifold Gaussian Conditioning}
\label{sec:appendix:gaussian_conditioning}
A Riemannian Gaussian Distribution with $\bm{\zeta} \sim \mathcal{N}(\bm{\mu}, \bm{\Sigma})$ can be decomposed into
\begin{gather}
	\bm{\zeta} = \begin{bmatrix}
		\bm{s}\\
		\bm{\xi}(\bm{s})
	\end{bmatrix}, \quad
	\bm{\mu} = \begin{bmatrix}
		\bm{\mu}^{I}\\
		\bm{\mu}^{O}
	\end{bmatrix}, \quad
	\bm{\Sigma} = \begin{bmatrix}
		\bm{\Sigma}^{II} & \bm{\Sigma}^{IO}\\
		\bm{\Sigma}^{OI} & \bm{\Sigma}^{OO}
	\end{bmatrix},
\end{gather}
to compute the probabilistic output $\bm{\xi}(\bm{s})$ given an input $\bm{s}$ through Gaussian conditioning.
$\bm{\xi}(\bm{s})$ is computed through the minimization problem (see, e.g., \cite{bishop2006pattern} for the Euclidean case)
\begin{align}
	J(\bm{\xi}(\bm{s})) &= \frac{1}{2} \mathrm{Log}_{\bm{\mu}}(\bm{\zeta})^\top \bm{\Sigma}^{-1} \mathrm{Log}_{\bm{\mu}}(\bm{\zeta})\\
	&= c + \mathrm{Log}_{\bm{\mu}^O}(\bm{\xi}(\bm{s}))^\top \bm{\Lambda}^{OI} \mathrm{Log}_{\bm{\mu}^I}(\bm{s}) +\nonumber\\
	&\mathrel{\phantom{=}} \frac{1}{2} \mathrm{Log}_{\bm{\mu}^O}(\bm{\xi}(\bm{s}))^\top \bm{\Lambda}^{OO} \mathrm{Log}_{\bm{\mu}^O}(\bm{\xi}(\bm{s}))\label{eq:error_function_gauss_cond}
\end{align}
where $c$ is a constant unaffected by $\bm{\xi}(\bm{s})$ and we have made use of the partitioning of the inverse of the covariance matrix
\begin{equation}
	\bm{\Lambda} = \bm{\Sigma}^{-1} = \begin{bmatrix}
		\bm{\Lambda}^{II} & \bm{\Lambda}^{IO}\\
		\bm{\Lambda}^{OI} & \bm{\Lambda}^{OO}
	\end{bmatrix}.
\end{equation}
A Taylor expansion of the minimization problem \eqref{eq:error_function_gauss_cond} yields
\begin{align}
	J(\bm{\xi}(\bm{s})) \approx c + \left(\mathrm{Log}_{\bm{\mu}^O}(\bm{\xi}(\bm{s})) + \Gamma_{O}(\bm{\epsilon})\right)^\top \bm{\Lambda}^{OI} \mathrm{Log}_{\bm{\mu}^I}(\bm{s}) +\nonumber\\
	\frac{1}{2} \left(\mathrm{Log}_{\bm{\mu}^O}(\bm{\xi}(\bm{s})) + \Gamma_{O}(\bm{\epsilon})\right)^\top \bm{\Lambda}^{OO} \left(\mathrm{Log}_{\bm{\mu}^O}(\bm{\xi}(\bm{s})) + \Gamma_{O}(\bm{\epsilon})\right) \label{eq:error_function_gauss_cond_taylor}
\end{align}
where $\Gamma_O(\bm{\epsilon})$ denotes the parallel transport $\Gamma_{\bm{\xi}(\bm{s})\rightarrow\bm{\mu}^O}(\bm{\epsilon})$, thus transporting the update $\bm{\epsilon}$ we want to apply to the current estimate of $\bm{\xi}(\bm{s})$ from the tangent space $\mathcal{T}_{\bm{\xi}(\bm{s})}\mathcal{M}$ to $\mathcal{T}_{\bm{\mu}^O}\mathcal{M}$.
Using parallel transport avoids the construction of an explicit adjoint operation on the manifold \cite{kim2014multvariate}, also note the similarity in formulation to the Lie group approach \cite{barfoot2014associating} for fusing Gaussians.
For the manifolds in this work, $\Gamma_{\bm{\xi}(\bm{s})\rightarrow\bm{\mu}^O}(\bm{\epsilon})$ can be formulated as left multiplication of a rotation matrix $\bm{R}_{\bm{\xi}(\bm{s})\rightarrow\bm{\mu}^O} \bm{\epsilon}$ which allows calculating the derivative of \eqref{eq:error_function_gauss_cond_taylor} w.r.t. $\bm{\epsilon}$.
In a Gauss-Newton optimization scheme, the next update is found by setting the derivative to zero, resulting in
\begin{align}
	\bm{R}_{\bm{\xi}(\bm{s})\rightarrow\bm{\mu}^O}^\top \bm{\Lambda}^{OO} \bm{R}_{\bm{\xi}(\bm{s})\rightarrow\bm{\mu}^O} \bm{\epsilon} &= -\bm{R}_{\bm{\xi}(\bm{s})\rightarrow\bm{\mu}^O}^\top \bm{\Lambda}^{OO} \mathrm{Log}_{\bm{\mu}^O}(\bm{\xi}(\bm{s}))\nonumber\\
	&\mathrel{\phantom{=}} -\bm{R}_{\bm{\xi}(\bm{s})\rightarrow\bm{\mu}^O}^\top \bm{\Lambda}^{OI} \mathrm{Log}_{\bm{\mu}^I}(\bm{s}).\label{eq:gauss_cond_derivative_zero}
\end{align}
Using the identity for the inverse of a partitioned matrix \cite{murphy2012machine}, it can be further simplified to
{\small
\begin{align}
	\bm{\epsilon} &= -\bm{R}_{\bm{\xi}(\bm{s})\rightarrow\bm{\mu}^O}^\top \mathrm{Log}_{\bm{\mu}^O}(\bm{\xi}(\bm{s})) - \bm{R}_{\bm{\xi}(\bm{s})\rightarrow\bm{\mu}^O}^\top \left(\bm{\Lambda}^{OO}\right)^{-1} \bm{\Lambda}^{OI} \mathrm{Log}_{\bm{\mu}^I}(\bm{s})\nonumber\\
	&= -\bm{R}_{\bm{\xi}(\bm{s})\rightarrow\bm{\mu}^O}^\top \mathrm{Log}_{\bm{\mu}^O}(\bm{\xi}(\bm{s})) + \bm{R}_{\bm{\xi}(\bm{s})\rightarrow\bm{\mu}^O}^\top \bm{\Sigma}^{OI} \left(\bm{\Sigma}^{II}\right)^{-1} \mathrm{Log}_{\bm{\mu}^I}(\bm{s})\label{eq:gauss_cond_derivative_zero_part_matrix}
\end{align}
}%
and is in line with the results of \cite{zeestraten2017manifold}.
Note that as the parallel transport is a rotation in our case, $\bm{R}_{\bm{\xi}(\bm{s})\rightarrow\bm{\mu}^O}^\top = \bm{R}_{\bm{\xi}(\bm{s})\rightarrow\bm{\mu}^O}^{-1}$ holds.
$\bm{\xi}(\bm{s})$ can then be computed iteratively through $	\bm{\xi}(\bm{s}) \leftarrow \mathrm{Exp}_{\bm{\xi}(\bm{s})}(\bm{\epsilon})$, after convergence, the covariance is computed to
\begin{align}
	\bm{\Sigma} &= \left(\bm{R}_{\bm{\xi}(\bm{s})\rightarrow\bm{\mu}^O}^\top \bm{\Lambda}^{OO} \bm{R}_{\bm{\xi}(\bm{s})\rightarrow\bm{\mu}^O}\right)^{-1}\label{eq:gauss_cond_cov}\\
	 &= \bm{R}_{\bm{\xi}(\bm{s})\rightarrow\bm{\mu}^O}^\top \left( \bm{\Sigma}^{OO} - \bm{\Sigma}^{OI} \left(\bm{\Sigma}^{II}\right)^{-1} \bm{\Sigma}^{IO} \right) \bm{R}_{\bm{\xi}(\bm{s})\rightarrow\bm{\mu}^O},\nonumber
\end{align}
again using the formulas for the inverse of a partitioned matrix and the inverse parallel transport $\bm{R}_{\bm{\xi}(\bm{s})\rightarrow\bm{\mu}^O}^\top$ to move the resulting covariance to $\mathcal{T}_{\bm{\xi}(\bm{s})} \mathcal{M} \otimes \mathcal{T}_{\bm{\xi}(\bm{s})} \mathcal{M}$.

\subsection{Movement Primitive Prediction through Conditioning}
\label{sec:appendix:prediction_conditioning}
Observing the similarities between \glspl{gp} with noisy observations \cite{williams2006gaussian} and \glspl{kmp}, we reformulate \glspl{kmp} to model a noisy function $\bm{\xi}(\bm{s}^*) = f(\bm{s}^*) + \bm{\delta}$ with process mean function value $\bm{e}$ and $\bm{\delta} \sim \mathcal{N} \left( \bm{0}, \bm{\Sigma}_{\bm{s}^*} \right)$.
The prior on these noisy observations then becomes $\mathrm{cov}(\bm{\xi}(\bm{s}^*)) = \bm{K} + \lambda\bm{\Sigma}$ where $\bm{\Sigma} = \mathrm{blockdiag}(\bm{\Sigma}_1, \bm{\Sigma}_2, \dots, \bm{\Sigma}_N)$ for the training data $\mathcal{D}$ (\Cref{sec:kmp}).
Note that similarly, parametric noise models have been used before \cite{murraysmith2001gaussian}.
This allows us to write
\begin{equation}
	\begin{bmatrix}
		\bm{\mu}\\
		\bm{\xi}(\bm{s}^*)
	\end{bmatrix} =
	\mathcal{N}\left(\bm{e}, \begin{bmatrix}
		\bm{K} + \lambda\bm{\Sigma} & \bm{k}^*\\
		{\bm{k}^*}^\top & \bm{k}(\bm{s}^*, \bm{s}^*)
	\end{bmatrix}\right)
\end{equation}
which can be solved similarly as before through minimizing
\begin{align}
	J(\bm{\xi}(\bm{s}^*)) &= c + \mathrm{Log}_{\bm{\xi}(\bm{s}^*)}(\bm{e})^\top \bm{\Lambda}^{OI} \mathrm{Log}_{\bm{\mu}}(\bm{e}) +\nonumber\\
	&\mathrel{\phantom{=}} \frac{1}{2} \mathrm{Log}_{\bm{\xi}(\bm{s}^*)}(\bm{e})^\top \bm{\Lambda}^{OO} \mathrm{Log}_{\bm{\xi}(\bm{s}^*)}(\bm{e}).\label{eq:error_function_gauss_cond_kmp}
\end{align}
Note that, unlike for Gaussian conditioning, we assume both $\bm{\Sigma}$ and the kernel function result to be in $\mathcal{T}_{\bm{\mu}} \mathcal{M} \otimes \mathcal{T}_{\bm{\mu}} \mathcal{M}$ and $\mathcal{T}_{\bm{\xi}(\bm{s}^*)} \mathcal{M} \otimes \mathcal{T}_{\bm{\xi}(\bm{s}^*)} \mathcal{M}$, respectively.
This holds trivially for $\bm{\Sigma}$ as the Riemannian Gaussian Distribution which is used as the probabilistic reference expresses its covariance in $\mathcal{T}_{\bm{\mu}} \mathcal{M} \otimes \mathcal{T}_{\bm{\mu}} \mathcal{M}$.
Due to this assumption, unlike the derivation in \Cref{sec:kmp:prediction} where basis functions were assumed to be expressed in $\mathcal{T}_{\bm{e}} \mathcal{M} \otimes \mathcal{T}_{\bm{e}} \mathcal{M}$, we do not require a parallel transport in the following equations.
Please see \Cref{sec:matrix_kernels} how this property is achieved for kernels.
The simplified Taylor expansion \eqref{eq:error_function_gauss_cond_taylor} then evaluates to
\begin{align}
	J(\bm{\xi}(\bm{s}^*)) &\approx c + \left(\mathrm{Log}_{\bm{\xi}(\bm{s}^*)}(\bm{e}) - \bm{\epsilon}\right)^\top \bm{\Lambda}^{OI} \mathrm{Log}_{\bm{\mu}}(\bm{e}) +\label{eq:error_function_kmp_taylor}\\
	&\mathrel{\phantom{\approx}} \frac{1}{2} \left(\mathrm{Log}_{\bm{\xi}(\bm{s}^*)}(\bm{e}) - \bm{\epsilon}\right)^\top \bm{\Lambda}^{OO} \left(\mathrm{Log}_{\bm{\xi}(\bm{s}^*)}(\bm{e}) - \bm{\epsilon}\right) \nonumber
\end{align}
as the update $\bm{\epsilon}$ does not require parallel transport.
Note the minus in front of $\bm{\epsilon}$, caused by $\bm{\epsilon}$ updating $\bm{\xi}(\bm{s}^*)$ which is the base of the $\mathrm{Log}$.
The update evaluates to
\begin{align}
	\bm{\epsilon} &= \left( \bm{\Lambda}^{OO} \right)^{-1} \left( \bm{\Lambda}^{OO} \mathrm{Log}_{\bm{\xi}(\bm{s}^*)}(\bm{e}) + \bm{\Lambda}^{OI} \mathrm{Log}_{\bm{\mu}}(\bm{e}) \right)\nonumber\\
	&= \mathrm{Log}_{\bm{\xi}(\bm{s}^*)}(\bm{e}) - \bm{k}^* \left( \bm{K} + \lambda \bm{\Sigma} \right)^{-1} \mathrm{Log}_{\bm{\mu}}(\bm{e}),
\end{align}
using the same relations as in \eqref{eq:gauss_cond_derivative_zero} and \eqref{eq:gauss_cond_derivative_zero_part_matrix} for Gaussian conditioning.
Note that when starting from an initial estimate $\bm{\xi}(\bm{s}^*) = \bm{e}$, convergence is given after one step through
\begin{equation}
	\bm{\xi}(\bm{s}^*) = \mathrm{Exp}_{\bm{e}}\left(- \bm{k}^* \left( \bm{K} + \lambda \bm{\Sigma} \right)^{-1} \mathrm{Log}_{\bm{\mu}}(\bm{e})\right).\label{eq:kmp_mean_prediction_alt}
\end{equation}
The covariance evaluates to
\begin{equation}
	\bm{\Sigma}(\bm{s}^*) = \left(\bm{\Lambda}^{OO}\right)^{-1} = \bm{k}(\bm{s}^*, \bm{s}^*) - \bm{k}^* \left(\bm{K} + \lambda\bm{\Sigma}\right)^{-1} \bm{k}^{*\top}	
\end{equation}
which is very similar to the result for Gaussian conditioning in \eqref{eq:gauss_cond_cov} but without parallel transport.

\section{Supplemental Derivations}

\subsection{Soft Null Space Adaptations in Riemannian KMPs}
\label{sec:appendix:soft_nullspace}
In this section, we derive the Riemannian null space prediction \eqref{eq:nullspace:prediction}.
Integrating the penalty term $\beta (\bm{\mu}_w - \hat{\bm{w}})^\top (\bm{\mu}_w - \hat{\bm{w}})$ to keep weights $\bm{\mu}_w$ close to desired ones $\hat{\bm{w}}$ in \eqref{eq:weight_space:costfcn}, we obtain
\begin{align}
	J(\bm{\mu}_w) & \approx \sum_{n=1}^N \left( \left( \bm{\Phi}_\parallel(\bm{s}_n)^\top \bm{\mu}_w + \Log_{\bm{\mu}_n} \bm{e} \right)^\top \bm{\Sigma}_n^{-1} \right. \nonumber\\
	& ~~~~~~~~~~ \left. \left( \bm{\Phi}_\parallel(\bm{s}_n)^\top \bm{\mu}_w + \Log_{\bm{\mu}_n} \bm{e} \right) \right)\nonumber\\
	& ~~~~~~~~~~ + \alpha \bm{\mu}_w^\top \bm{\mu}_w + \beta (\bm{\mu}_w - \hat{\bm{w}})^\top (\bm{\mu}_w - \hat{\bm{w}}).\label{eq:nullspace:costfcn}
\end{align}
Setting the derivative of this cost function to zero, we obtain
\begin{align}
	\left( \bm{\Phi}_\parallel(\bm{s}) \bm{\Sigma}^{-1} \bm{\Phi}_\parallel^\top(\bm{s}) + \left( \alpha + \beta \right) \bm{I} \right) \bm{\mu}_w^* =\nonumber\\
	-\bm{\Phi}_\parallel(\bm{s}) \bm{\Sigma}^{-1} \Log_{\bm{\mu}}(\bm{e}) + \beta \hat{\bm{w}}\label{eq:nullspace:derivative}
\end{align}
Defining $\lambda = \alpha + \beta$, this is the same result as \eqref{eq:weight_space_derivative} except for the additional term $\beta \hat{\bm{w}}$.
As \eqref{eq:nullspace:derivative} is similar to the Euclidean derivations of \cite{silverio2023nonparametric}, we can thus follow the steps laid out therein and in \Cref{sec:kmp:prediction}, arriving at the optimal solution
\begin{align}
	\bm{\mu}_w^* &= -\bm{\Phi}_\parallel(\bm{s}) \left( \bm{\Phi}_\parallel(\bm{s})^\top \bm{\Phi}_\parallel(\bm{s}) + \lambda \bm{\Sigma} \right)^{-1} \Log_{\bm{\mu}}(\bm{e})\nonumber\\
	&~~+\frac{\beta}{\lambda}\left[ \bm{I} - \bm{\Phi}_\parallel(\bm{s}) \left( \bm{\Phi}_\parallel(\bm{s})^\top \bm{\Phi}_\parallel(\bm{s}) + \lambda \bm{\Sigma} \right)^{-1} \bm{\Phi}_\parallel^\top(\bm{s}) \right] \hat{\bm{w}}.\nonumber
\end{align}
From the desired modification $\hat{\bm{\xi}} = \Exp_{\bm{e}} \left( \hat{\bm{\Phi}}_\parallel^\top \hat{\bm{w}} \right)$ resulting in weights $\hat{\bm{w}} = \hat{\bm{\Phi}}_\parallel \left(\hat{\bm{\Phi}}_\parallel^\top \hat{\bm{\Phi}}_\parallel\right)^{-1} \Gamma_{\bm{e} \rightarrow \bm{\xi}(\bm{s}^*)} \left(\Log_{\bm{\xi}(\bm{s}^*)}(\hat{\bm{\xi}})\right)$, we kernelize the solution to obtain \eqref{eq:nullspace:prediction}, assuming $\frac{\beta}{\lambda} = 1$ without loss of generality -- this factor can, e.g., be pulled into $\hat{\bm{w}}$.

\subsection{Restoring the Lower-triangular Form of \texorpdfstring{$\bm{L}$}{L}}
\label{sec:appendix:rediag}
\begin{algorithm}[t]
\caption{\label{algo:ckmp:remove_measurements} Removing measurements in a cKMP.}
\begin{algorithmic}
\For{$i$ in \textit{measurements to remove}}
	\State $\bm{s} \gets [\bm{s} \setminus \bm{s}_i]$, $\bm{\mu} \gets [\bm{\mu} \setminus \bm{\mu}_i]$, $\bm{\Sigma} \gets [\bm{\Sigma} \setminus \bm{\Sigma}_i]$ \Comment{remove}
	\State $\bm{K} \gets [\bm{K} \setminus \bm{K}_{i, j}]$ \Comment{remove}
\EndFor
\State $\bm{L} \gets \bm{P} \cdot \bm{L}$ \Comment{reorder using $\bm{P}$}
\While{$\bm{L}$ not lower-triangular}
	\State $i, j \gets \mathrm{max}(j-i | \bm{L}_{ij} \neq 0)$ \Comment{Furthest off-diagonal}
	\State $\bm{L} \gets \bm{L} \cdot \bm{G}(i,j,\theta)$ \Comment{Zero out $\bm{L}_{ij}$ using \eqref{eq:givens_rotation} -- \eqref{eq:givens_rotation_angle}}
\EndWhile
\end{algorithmic}
\end{algorithm}
When removing a measurement (\Cref{sec:computational_improvements:removing}) in the Cholesky-factorized \gls{kmp} (cKMP) before the end, the decomposition $\bm{L}$ needs reordering.
While still a valid decomposition, the permuted\footnote{Note that the permutation $\bm{P} \bm{L}$ can be implemented by reordering the matrix, not requiring $\mathcal{O}(N^3)$ matrix multiplications.} $\bm{L}^{'} = \bm{P} \bm{L}$ is not lower-triangular anymore in case we want to remove one or more measurements before the last measurement.
Instead, it has rows ``sticking out'', e.g.,
\begin{equation}
\tiny
	\bm{L}^{'} = \begin{pmatrix}
		l_{1,1} & \hdots & 0 & \hdots & 0\\
		\hdots & \ddots & 0 & \hdots & 0\\
		l_{m,1} & \hdots & l_{m,m} & l_{m,m+1} & 0\\
		\hdots & \hdots & \hdots & \ddots & 0\\
		l_{n,1} & \hdots & \hdots & l_{n,n-1} & 0
	\end{pmatrix},
\end{equation}
where entries $l_{m, m+1}$ (or in general, $l_{m, m+k}$) extend beyond the lower triangle.
After removing the last rows and columns, $\bm{L}^{'}$ is unusable for forward- / backward-substitution as it would result in an overdetermined system of equations.
Entries reaching over the diagonal can be zeroed out using Givens rotations ($c_\theta = \mathrm{cos}(\theta)$, $s_\theta = \mathrm{sin}(\theta)$) \cite[Ch. 12.5]{golub1996matrix}
\begin{equation}
\tiny
	\bm{G}(i,j,\theta) = \begin{pmatrix}
		1 & \dots & 0 & 0 & \dots & 0\\
		\vdots & \ddots & \vdots & \vdots & & \vdots\\
		0 & \dots & c_\theta & s_\theta & \dots & 0\\
		0 & \dots & -s_\theta & c_\theta & \dots & 0\\
		\vdots & & \vdots & \vdots & \ddots & \vdots\\
		1 & \dots & 0 & 0 & \dots & 0
	\end{pmatrix}.
	\label{eq:givens_rotation}
\end{equation}
When applied on the right side, i.e. $\bm{L}^* = \bm{L} \cdot \bm{G}(i,j,\theta)$, they act on the row $\bm{L}_i = \left( *, \dots, *, \bm{L}_{i,j-1}, \bm{L}_{i,j}, *, \dots, * \right)$, leading to an equation for the element we would like to zero out
\begin{align}
	\bm{L}_{i,j}^* = -s_\theta \bm{L}_{i,j-1} + c_\theta \bm{L}_{i,j} \stackrel{!}{=} 0.
\end{align}
We can thus determine the entries of the Givens rotation to
\begin{gather}
	r = \sqrt{\bm{L}_{i,j-1}^2 + \bm{L}_{i,j}^2}, \quad c_\theta = \frac{\bm{L}_{i,j-1}}{r}, \quad s_\theta = \frac{\bm{L}_{i,j}}{r},
	\label{eq:givens_rotation_angle}
\end{gather}
where we have used the trigonometric identity $c_\theta^2 + s_\theta^2 = 1$.
These rotations have to be applied to all row vectors in rows $\left[i, \hdots, n\right]$; we can omit applying it to the last rows we want to remove.
Sorting rows to be processed in order $\mathrm{max} \left( k \right); \mathrm{min} \left( i \right)$, where $k$ is the number of elements the row reaches over the diagonal, we can restore the lower triangular shape of $\bm{L}^{'}$.
The full algorithm is outlined in \Cref{algo:ckmp:remove_measurements}.
The runtime depends on which measurement we want to remove, as the removed and all following rows have to be partially recomputed -- removing towards the end is thus progressively cheaper.
As mentioned in \Cref{sec:computational_improvements:removing}, for removing measurements at the end we can simply drop rows, making the Cholesky decomposition especially appealing for continuously updating via points as shown in \Cref{sec:eval:adapting}.
\end{appendices}
\end{document}

%% file: preamble.tex
\usepackage{amsmath,amssymb,amsfonts}
\usepackage{algorithmicx}
\usepackage[pdftex]{graphicx} 
\usepackage{textcomp} 
\usepackage{xcolor}
\def\BibTeX{{\rm B\kern-.05em{\sc i\kern-.025em b}\kern-.08em
    T\kern-.1667em\lower.7ex\hbox{E}\kern-.125emX}}

\usepackage{hyperref}

\usepackage{cleveref}
\Crefname{equation}{Eq.}{Eqs.} 
\Crefname{figure}{Fig.}{Figs.}

\graphicspath{{images-src/}{images-bin/}} 

\usepackage[textsize=scriptsize]{todonotes} 
\newcommand{\annotategraphicsmulti}[3][]{
  \begin{tikzpicture}[%
  every node/.style={draw=black, black, text opacity=1, fill=white, fill opacity=0.75,inner sep=0.5mm, #1},%
  ]
  \node[anchor=south west,inner sep=0, draw=none] (image) at (0,0) {
    #2
  };
  \begin{scope}[x={(image.south east)},y={(image.north west)}]
    #3
  \end{scope}
  \end{tikzpicture}%
}

\newcommand*\lref[1]{\tikz[baseline=(char.base)]{\node[shape=rectangle,draw,inner sep=2pt] (char) {\scriptsize #1};}}

%% file: style/dlr_colors.tex
\definecolor{dlrprim1}{HTML}{000000} 
\definecolor{dlrprim2}{HTML}{666666} 
\definecolor{dlrprim3}{HTML}{b9cad2}
\definecolor{dlrprim4}{HTML}{ffffff} 

\definecolor{dlrblue1}{HTML}{00658b} 
\definecolor{dlrblue2}{HTML}{3b98cb}
\definecolor{dlrblue3}{HTML}{6cb9dc}
\definecolor{dlrblue4}{HTML}{a7d3ec}
\definecolor{dlrblue5}{HTML}{d1e8fa}

\definecolor{dlryellow1}{HTML}{d2ae3d}  
\definecolor{dlryellow2}{HTML}{f2cd51} 
\definecolor{dlryellow3}{HTML}{f8de53}
\definecolor{dlryellow3}{HTML}{fcea7a}
\definecolor{dlryellow3}{HTML}{fff8be}

\definecolor{dlrgreen1}{HTML}{82a043} 
\definecolor{dlrgreen2}{HTML}{a6bf51}
\definecolor{dlrgreen3}{HTML}{cad55c}
\definecolor{dlrgreen4}{HTML}{d9df78}
\definecolor{dlrgreen5}{HTML}{e6eaaf}

\definecolor{dlrgray1}{HTML}{666666} 
\definecolor{dlrgray2}{HTML}{868585}
\definecolor{dlrgray3}{HTML}{b1b1b1}
\definecolor{dlrgray4}{HTML}{cfcfcf}
\definecolor{dlrgray5}{HTML}{ebebeb}

\definecolor{lavender}{HTML}{e6e6fA}
\definecolor{pastelpurple}{HTML}{c3b1e1}
\definecolor{deeppastelpurple}{HTML}{58378f}

%% file: main.bbl
\begin{thebibliography}{41}
\providecommand{\natexlab}[1]{#1}
\providecommand{\url}[1]{#1}
\csname url@samestyle\endcsname
\providecommand{\newblock}{\relax}
\providecommand{\bibinfo}[2]{#2}
\providecommand{\BIBentrySTDinterwordspacing}{\spaceskip=0pt\relax}
\providecommand{\BIBentryALTinterwordstretchfactor}{4}
\providecommand{\BIBentryALTinterwordspacing}{\spaceskip=\fontdimen2\font plus
\BIBentryALTinterwordstretchfactor\fontdimen3\font minus
  \fontdimen4\font\relax}
\providecommand{\BIBforeignlanguage}[2]{{%
\expandafter\ifx\csname l@#1\endcsname\relax
\typeout{** WARNING: IEEEtranN.bst: No hyphenation pattern has been}%
\typeout{** loaded for the language `#1'. Using the pattern for}%
\typeout{** the default language instead.}%
\else
\language=\csname l@#1\endcsname
\fi
#2}}
\providecommand{\BIBdecl}{\relax}
\BIBdecl

\bibitem[Mühlbauer et~al.(2025{\natexlab{a}})Mühlbauer, Chalon, Silvério,
  and Albu-Schäffer]{muehlbauer2025control}
M.~Mühlbauer, M.~Chalon, J.~Silvério, and A.~Albu-Schäffer, ``Control system
  for a torque-sensitive space robotic arm,'' in \emph{76th International
  Astronautical Congress, IAC 2025}, ser. IAC 2025.\hskip 1em plus 0.5em minus
  0.4em\relax International Astronautical Federation (IAF), 2025, pp. 310--321.

\bibitem[Schmidt et~al.(2025)Schmidt, Höhn, Singh, Walter, Mühlbauer, Pap,
  Panzirsch, Lodewikus, Van~Strien, Zhang, Dammann, Broghammer, Hahn, Noack,
  Lii, Albu-Schäffer, and Hulin]{schmidt2025conceptual}
A.~Schmidt, D.~Höhn, H.~Singh, M.~Walter, M.~Mühlbauer, E.-U. Pap,
  M.~Panzirsch, B.~Lodewikus, H.~Van~Strien, S.~Zhang, A.~Dammann,
  F.~Broghammer, A.~Hahn, T.~Noack, N.~Y. Lii, A.~Albu-Schäffer, and T.~Hulin,
  ``Conceptual framework for teleoperated robotic recovery of large subsea
  {UXOs},'' in \emph{Conference on Telepresence}.\hskip 1em plus 0.5em minus
  0.4em\relax IEEE, 2025, pp. 197--202.

\bibitem[Quere et~al.(2024)Quere, Stulp, Filliat, and
  Silvério]{quere2024probabilistic}
G.~Quere, F.~Stulp, D.~Filliat, and J.~Silvério, ``A probabilistic approach
  for learning and adapting shared control skills with the human in the loop,''
  in \emph{2024 IEEE International Conference on Robotics and Automation
  (ICRA)}.\hskip 1em plus 0.5em minus 0.4em\relax IEEE, 05 2024, pp.
  15\,728--15\,734.

\bibitem[Zeestraten et~al.(2017)Zeestraten, Havoutis, Silvério, Calinon, and
  Caldwell]{zeestraten2017manifold}
M.~J.~A. Zeestraten, I.~Havoutis, J.~Silvério, S.~Calinon, and D.~G. Caldwell,
  ``An approach for imitation learning on {R}iemannian manifolds,'' \emph{IEEE
  Robotics and Automation Letters}, vol.~2, no.~3, pp. 1240--1247, 2017.

\bibitem[Calinon(2020)]{calinon2020gaussians}
S.~Calinon, ``Gaussians on {Riemannian} manifolds: Applications for robot
  learning and adaptive control,'' \emph{IEEE Robotics \& Automation Magazine},
  vol.~27, no.~2, pp. 33--45, 2020.

\bibitem[Williams and Rasmussen(2006)]{williams2006gaussian}
C.~K.~I. Williams and C.~E. Rasmussen, \emph{Gaussian processes for machine
  learning}.\hskip 1em plus 0.5em minus 0.4em\relax MIT press Cambridge, MA,
  2006.

\bibitem[Hüllermeier and Waegeman(2021)]{huellermeier2021aleatoric}
E.~Hüllermeier and W.~Waegeman, ``Aleatoric and epistemic uncertainty in
  machine learning: an introduction to concepts and methods,'' \emph{Machine
  Learning}, vol. 110, no.~3, pp. 457--506, 2021.

\bibitem[Huang et~al.(2019)Huang, Rozo, Silvério, and
  Caldwell]{huang2019kernelized}
Y.~Huang, L.~Rozo, J.~Silvério, and D.~G. Caldwell, ``Kernelized movement
  primitives,'' \emph{The International Journal of Robotics Research}, vol.~38,
  no.~7, pp. 833--852, 2019.

\bibitem[Huang et~al.(2021)Huang, Abu-Dakka, Silvério, and
  Caldwell]{huang2021toward}
Y.~Huang, F.~J. Abu-Dakka, J.~Silvério, and D.~G. Caldwell, ``Toward
  orientation learning and adaptation in cartesian space,'' \emph{IEEE
  Transactions on Robotics}, vol.~37, no.~1, pp. 82--98, 2021.

\bibitem[Silvério and Huang(2023)]{silverio2023nonparametric}
J.~Silvério and Y.~Huang, ``A non-parametric skill representation with soft
  null space projectors for fast generalization,'' in \emph{2023 IEEE
  International Conference on Robotics and Automation (ICRA)}, 2023, pp.
  2988--2994.

\bibitem[Knauer et~al.(2025)Knauer, Albu-Schäffer, Stulp, and
  Silvério]{knauer2025interactive}
M.~Knauer, A.~Albu-Schäffer, F.~Stulp, and J.~Silvério, ``Interactive
  incremental learning of generalizable skills with local trajectory
  modulation,'' \emph{IEEE Robotics and Automation Letters}, vol.~10, no.~4,
  pp. 3398--3405, 2025.

\bibitem[Mühlbauer et~al.(2025{\natexlab{b}})Mühlbauer, Weber, Calinon,
  Stulp, Albu-Schäffer, and Silvério]{muehlbauer2025unified}
\BIBentryALTinterwordspacing
M.~Mühlbauer, B.~Weber, S.~Calinon, F.~Stulp, A.~Albu-Schäffer, and
  J.~Silvério, ``A unified framework for probabilistic dynamic-, trajectory-
  and vision-based virtual fixtures,'' 2025. [Online]. Available:
  \url{https://arxiv.org/abs/2506.10239}
\BIBentrySTDinterwordspacing

\bibitem[Mühlbauer et~al.(2024)Mühlbauer, Hulin, Weber, Calinon, Stulp,
  Albu-Schäffer, and Silvério]{muehlbauer2024probabilistic}
M.~Mühlbauer, T.~Hulin, B.~Weber, S.~Calinon, F.~Stulp, A.~Albu-Schäffer, and
  J.~Silvério, ``A probabilistic approach to multi-modal adaptive virtual
  fixtures,'' \emph{IEEE Robotics and Automation Letters}, vol.~9, no.~6, pp.
  5298--5305, 2024.

\bibitem[Padalkar et~al.(2025)Padalkar, Stulp, Neumann, and
  Silvério]{padalkar2025towards}
A.~Padalkar, F.~Stulp, G.~Neumann, and J.~Silvério, ``Towards safe and
  efficient learning in the wild: Guiding {RL} with constrained
  uncertainty-aware movement primitives,'' \emph{IEEE Robotics and Automation
  Letters}, vol.~10, no.~7, pp. 6880--6887, 2025.

\bibitem[Jaquier et~al.(2024)Jaquier, Rozo, and Asfour]{jaquier2024unraveling}
N.~Jaquier, L.~Rozo, and T.~Asfour, ``Unraveling the single tangent space
  fallacy: An analysis and clarification for applying {Riemannian} geometry in
  robot learning,'' in \emph{2024 IEEE International Conference on Robotics and
  Automation (ICRA)}.\hskip 1em plus 0.5em minus 0.4em\relax IEEE, 2024, pp.
  242--249.

\bibitem[Li et~al.(2025)Li, Xu, Wang, Ye, Chen, Song, and
  Huang]{li2025orientation}
\BIBentryALTinterwordspacing
G.~Li, P.~Xu, R.~Wang, Q.~Ye, J.~Chen, D.~Song, and Y.~Huang, ``Orientation
  learning and adaptation towards simultaneous incorporation of multiple local
  constraints,'' 2025. [Online]. Available:
  \url{https://arxiv.org/abs/2510.07986}
\BIBentrySTDinterwordspacing

\bibitem[Lopez-Custodio et~al.(2024)Lopez-Custodio, Bharath, Kucukyilmaz, and
  Preston]{lopez2023non}
\BIBentryALTinterwordspacing
P.~C. Lopez-Custodio, K.~Bharath, A.~Kucukyilmaz, and S.~P. Preston,
  ``Non-parametric regression for robot learning on manifolds,'' 2024.
  [Online]. Available: \url{https://arxiv.org/abs/2310.19561}
\BIBentrySTDinterwordspacing

\bibitem[Álvarez et~al.(2012)Álvarez, Rosasco, and
  Lawrence]{alvarez2012kernels}
M.~A. Álvarez, L.~Rosasco, and N.~D. Lawrence, ``Kernels for vector-valued
  functions: A review,'' \emph{Foundations and Trends® in Machine Learning},
  vol.~4, no.~3, pp. 195--266, 2012.

\bibitem[Micheli and Glaunès(2013)]{micheli2013matrix}
\BIBentryALTinterwordspacing
M.~Micheli and J.~A. Glaunès, ``Matrix-valued kernels for shape deformation
  analysis,'' 2013. [Online]. Available: \url{https://arxiv.org/abs/1308.5739}
\BIBentrySTDinterwordspacing

\bibitem[Wang et~al.(2026)Wang, Wu, Liu, Zhang, Silv{\'e}rio, and
  Huang]{wang2026computationally}
Y.~Wang, S.~Wu, C.~Liu, C.~Zhang, J.~Silv{\'e}rio, and Y.~Huang, ``A
  computationally efficient nonparametric approach for robot imitation
  learning,'' in \emph{Proceedings of 2026 IEEE International Conference on
  Robotics and Automation (ICRA)}, 2026.

\bibitem[Ti et~al.(2023)Ti, Razmjoo, Gao, Zhao, and Calinon]{ti2023geometric}
B.~Ti, A.~Razmjoo, Y.~Gao, J.~Zhao, and S.~Calinon, ``A geometric optimal
  control approach for imitation and generalization of manipulation skills,''
  \emph{Robotics and Autonomous Systems}, vol. 164, p. 104413, 2023.

\bibitem[Thomas~Fletcher(2013)]{fletcher2012geodesic}
P.~Thomas~Fletcher, ``Geodesic regression and the theory of least squares on
  {R}iemannian manifolds,'' \emph{International Journal of Computer Vision},
  vol. 105, no.~2, pp. 171--185, 2013.

\bibitem[Kim et~al.(2014)Kim, Adluru, Collins, Chung, Bendin, Johnson,
  Davidson, and Singh]{kim2014multvariate}
H.~J. Kim, N.~Adluru, M.~D. Collins, M.~K. Chung, B.~B. Bendin, S.~C. Johnson,
  R.~J. Davidson, and V.~Singh, ``Multivariate general linear models ({MGLM})
  on {R}iemannian manifolds with applications to statistical analysis of
  diffusion weighted images,'' in \emph{2014 IEEE Conference on Computer Vision
  and Pattern Recognition}.\hskip 1em plus 0.5em minus 0.4em\relax IEEE, 2014,
  pp. 2705--2712.

\bibitem[Calinon(2015)]{calinon2015tutorial}
S.~Calinon, ``A tutorial on task-parameterized movement learning and
  retrieval,'' \emph{Intelligent Service Robotics}, vol.~9, no.~1, pp. 1--29,
  Sep 2015.

\bibitem[Mallasto and Feragen(2018)]{mallasto2018wrapped}
A.~Mallasto and A.~Feragen, ``Wrapped {Gaussian} process regression on
  {Riemannian} manifolds,'' in \emph{Proceedings of the IEEE Conference on
  Computer Vision and Pattern Recognition (CVPR)}, June 2018.

\bibitem[Bishop(2006)]{bishop2006pattern}
C.~M. Bishop, \emph{Pattern Recognition and Machine Learning}, ser. Information
  Science and Statistics.\hskip 1em plus 0.5em minus 0.4em\relax Springer,
  2006.

\bibitem[M{\"u}hlbauer et~al.(2026{\natexlab{a}})M{\"u}hlbauer, Sachtler,
  Albu-Sch{\"a}ffer, and Silv{\'e}rio]{muehlbauer2026geometric}
M.~M{\"u}hlbauer, A.~Sachtler, A.~O. Albu-Sch{\"a}ffer, and J.~Silv{\'e}rio,
  ``Geometric and model priors in motion primitives,'' in \emph{2026 IEEE
  International Conference on Robotics and Automation, Workshop on Geometry in
  the Age of Data‑Driven Robotics}, 2026.

\bibitem[Feragen et~al.(2015)Feragen, Lauze, and Hauberg]{feragen2015geodesic}
A.~Feragen, F.~Lauze, and S.~Hauberg, ``Geodesic exponential kernels: When
  curvature and linearity conflict,'' in \emph{2015 IEEE Conference on Computer
  Vision and Pattern Recognition (CVPR)}.\hskip 1em plus 0.5em minus
  0.4em\relax IEEE, 2015, pp. 3032--3042.

\bibitem[Feragen and Hauberg(2016)]{feragen2016open}
\BIBentryALTinterwordspacing
A.~Feragen and S.~Hauberg, ``Open problem: Kernel methods on manifolds and
  metric spaces. what is the probability of a positive definite geodesic
  exponential kernel?'' in \emph{29th Annual Conference on Learning Theory},
  ser. Proceedings of Machine Learning Research, vol.~49.\hskip 1em plus 0.5em
  minus 0.4em\relax Columbia University, New York, New York, USA: PMLR, 23--26
  Jun 2016, pp. 1647--1650. [Online]. Available:
  \url{https://proceedings.mlr.press/v49/feragen16.html}
\BIBentrySTDinterwordspacing

\bibitem[Jaquier et~al.(2020)Jaquier, Rozo, Calinon, and
  B\"urger]{jaquier2020bayesian}
\BIBentryALTinterwordspacing
N.~Jaquier, L.~Rozo, S.~Calinon, and M.~B\"urger, ``Bayesian optimization meets
  {Riemannian} manifolds in robot learning,'' in \emph{Proceedings of the
  Conference on Robot Learning}, ser. Proceedings of Machine Learning Research,
  vol. 100.\hskip 1em plus 0.5em minus 0.4em\relax PMLR, 30 Oct--01 Nov 2020,
  pp. 233--246. [Online]. Available:
  \url{https://proceedings.mlr.press/v100/jaquier20a.html}
\BIBentrySTDinterwordspacing

\bibitem[Jaquier et~al.(2022)Jaquier, Borovitskiy, Smolensky, Terenin, Asfour,
  and Rozo]{jacquier2022geometry}
N.~Jaquier, V.~Borovitskiy, A.~Smolensky, A.~Terenin, T.~Asfour, and L.~Rozo,
  ``Geometry-aware bayesian optimization in robotics using {Riemannian}
  {Matérn} kernels,'' in \emph{Proceedings of the 5th Conference on Robot
  Learning}, ser. Proceedings of Machine Learning Research, vol. 164.\hskip 1em
  plus 0.5em minus 0.4em\relax PMLR, 08--11 Nov 2022, pp. 794--805.

\bibitem[Caponnetto et~al.(2008)Caponnetto, Micchelli, Pontil, and
  Ying]{caponnetto2008universal}
A.~Caponnetto, C.~A. Micchelli, M.~Pontil, and Y.~Ying, ``Universal multi-task
  kernels,'' \emph{The Journal of Machine Learning Research}, vol.~9, pp.
  1615--1646, 2008.

\bibitem[Carmeli et~al.(2010)Carmeli, de~Vito, Toigo, and
  Umanit{\'a}]{carmeli2010vector}
C.~Carmeli, E.~de~Vito, A.~Toigo, and V.~Umanit{\'a}, ``Vector valued
  reproducing kernel hilbert spaces and universality,'' \emph{Analysis and
  Applications}, vol.~08, no.~01, pp. 19--61, 2010.

\bibitem[Murphy(2012)]{murphy2012machine}
K.~P. Murphy, \emph{Machine learning: a probabilistic perspective}.\hskip 1em
  plus 0.5em minus 0.4em\relax MIT Press, 2012.

\bibitem[Osborne(2010)]{osborne2010bayesian}
M.~A. Osborne, ``Bayesian {Gaussian} processes for sequential prediction,
  optimisation and quadrature,'' Ph.D. dissertation, Oxford University, UK,
  2010.

\bibitem[M{\"u}hlbauer et~al.(2026{\natexlab{b}})M{\"u}hlbauer, Werner,
  Balachandran, Hulin, Silv{\'e}rio, Stulp, and
  Albu-Sch{\"a}ffer]{muehlbauer2026stabilizingarxiv}
\BIBentryALTinterwordspacing
M.~M{\"u}hlbauer, N.~Werner, R.~Balachandran, T.~Hulin, J.~Silv{\'e}rio,
  F.~Stulp, and A.~Albu-Sch{\"a}ffer, ``Passive variable impedance for shared
  control,'' \emph{arXiv preprint arXiv:2604.20557}, 2026. [Online]. Available:
  \url{https://arxiv.org/abs/2604.20557}
\BIBentrySTDinterwordspacing

\bibitem[Knauer et~al.(2026{\natexlab{a}})Knauer, Bustamante, Eiband,
  Albu-Schaffer, Stulp, and Silverio]{knauer2026irosa}
M.~Knauer, S.~Bustamante, T.~Eiband, A.~Albu-Schaffer, F.~Stulp, and
  J.~Silverio, ``Irosa: Interactive robot skill adaptation using natural
  language,'' \emph{IEEE Robotics and Automation Letters}, vol.~11, no.~4, pp.
  5246--5253, 2026.

\bibitem[Knauer et~al.(2026{\natexlab{b}})Knauer, Gieraths, Mai, Bustamante,
  Albu-Schäffer, Stulp, and Silvério]{knauer2026clasp}
\BIBentryALTinterwordspacing
M.~Knauer, V.~Gieraths, T.~Mai, S.~Bustamante, A.~Albu-Schäffer, F.~Stulp, and
  J.~Silvério, ``Clasp: Language-driven robot skill selection and composition
  using task-parameterized learning,'' 2026. [Online]. Available:
  \url{https://arxiv.org/abs/2606.08169}
\BIBentrySTDinterwordspacing

\bibitem[Barfoot and Furgale(2014)]{barfoot2014associating}
T.~D. Barfoot and P.~T. Furgale, ``Associating uncertainty with
  three-dimensional poses for use in estimation problems,'' \emph{IEEE
  Transactions on Robotics}, vol.~30, no.~3, pp. 679--693, 2014.

\bibitem[Murray-Smith and Girard(2001)]{murraysmith2001gaussian}
R.~Murray-Smith and A.~Girard, ``Gaussian process priors with arma noise
  models,'' in \emph{Irish Signals and Systems Conference}, 2001.

\bibitem[Golub and Van~Loan(1996)]{golub1996matrix}
G.~H. Golub and C.~F. Van~Loan, \emph{Matrix computations (3rd ed.)}.\hskip 1em
  plus 0.5em minus 0.4em\relax Johns Hopkins University Press, 1996.

\end{thebibliography}
